\documentclass{article} 
\usepackage{iclr2027_conference,times}

\usepackage{amsmath,amsfonts,bm}

\def\eqref#1{equation~\ref{#1}}

\def\plaineqref#1{\ref{#1}}

\def\1{\bm{1}}

\DeclareMathAlphabet{\mathsfit}{\encodingdefault}{\sfdefault}{m}{sl}
\SetMathAlphabet{\mathsfit}{bold}{\encodingdefault}{\sfdefault}{bx}{n}

\usepackage{graphicx}
\usepackage{booktabs}
\usepackage{hyperref}
\usepackage{url}

\usepackage{wrapfig}
\usepackage{pifont}
\usepackage{capt-of}
\usepackage{multirow}
\newcommand{\cmark}{\ding{51}}
\newcommand{\xmark}{\ding{55}}

\title{NHO: A Neural Hamiltonian Operator for Anchor-based Region Localization and Dense Correspondence}

\author{
Jing Li$^{1}$, Yawei Luo$^{2}$\thanks{Corresponding authors.},
Xiangze Meng$^{1}$, Ying Li$^{3}$, Tieru Wu$^{1}$,
Rui Ma$^{1}$\footnotemark[1] \\
$^{1}$Jilin University, $^{2}$Zhejiang University, $^{3}$North China University of Technology
}

\iclrfinalcopy 
\begin{document}

\maketitle

\lhead{Preprint}

\begin{figure}[ht]
    \centering
    \includegraphics[width=\linewidth]{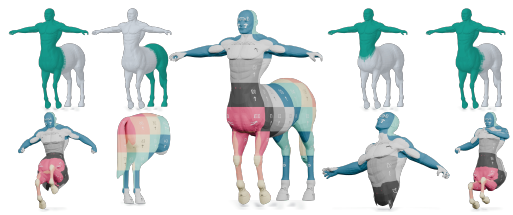}
    \caption{\textbf{NHO (Neural Hamiltonian Operator)} models a neural eigenspace using spatial cues, supporting both region localization and dense correspondence estimation for non-rigid partial-to-full shape matching.}
    \label{fig:teaser}
\end{figure}

\begin{abstract}

Non-rigid partial-to-full shape correspondence from sparse anchors requires identifying the corresponding region on the full surface and recovering dense correspondences between the partial shape and that region. We present NHO, which combines sparse anchors with the intrinsic geometry of the partial shape to learn a neural Hamiltonian operator whose localized eigenspace encodes both the region support and intrinsic coordinates for dense correspondence. NHO parameterizes the Hamiltonian potential as an intrinsic neural field and optimizes it using anchor evidence together with spectral and geometric constraints. To resolve the spatial ambiguity left by sparse anchors, we introduce reciprocal refinement between operator estimation and correspondence recovery. At each round, the current eigenspace provides spectral coordinates and restricts matching to its induced support, while geometrically reliable correspondences provide additional evidence for updating the potential. After refinement, aggregated eigenfunction energy yields the final localization, and the recovered map initializes dense correspondence refinement. Experiments demonstrate competitive accuracy on both tasks and robustness to uniform scaling and rotation.


\end{abstract}

\section{Introduction}
\label{intro}
Establishing dense correspondence between non-rigid partial and full shapes is a fundamental problem in 3D shape analysis, with broad applications in shape retrieval, animation, and geometric learning. Compared with full-to-full matching, partial-to-full correspondence is inherently more ambiguous: only an unknown subset of the full shape admits valid correspondences, while the remaining geometry introduces structured distractors. A small number of reliable correspondences can often be obtained from manual landmarks or confident descriptor matches, providing a practical starting point for addressing this problem~\citep{rakprayoon2021part,bensaid2023partial}.

However, sparse correspondences provide spatial constraints at only a few locations. The extent of the corresponding region and the correspondences for the remaining points are still unknown. These two estimation problems are coupled. Identifying the region excludes irrelevant geometry and reduces correspondence ambiguity, while additional correspondences provide spatial evidence for determining the region. With limited spatial cues, uncertainty in either estimate can propagate to the other.

Existing methods differ in which aspect of this coupled inference they emphasize. Correspondence-oriented methods~\citep{rakprayoon2021part,attaiki2021dpfm,cao2023unsupervised} recover dense pointwise correspondences either by expanding sparse initial matches or by estimating functional maps from learned vertex-wise features. Localization-oriented methods~\citep{rampini2019correspondence,bensaid2022partial,bensaid2023partial} focus on the other unknown: they represent partiality as a region indicator or a spatially varying potential and recover the corresponding support through neural interpolation or spectral alignment. Joint spectral formulations~\citep{rodola2017partial,litany2017fully,postolache2020parametric,wu2020partial} explicitly connect the two by jointly estimating partial support and a functional map, or by constructing spectral bases localized in the latent corresponding region. When sparse correspondences are available, existing methods primarily use them to initialize correspondence refinement or to fit a partiality indicator. How to lift such sparse evidence into a local spectral representation that supports both region localization and dense correspondence recovery remains underexplored. App.~\ref{app:method_comparison} provides a comparison with representative methods.

In this work, we introduce NHO, a method for learning a \textbf{neural Hamiltonian operator} on the full surface by parameterizing its potential as an intrinsic neural field. Sparse correspondences serve as anchors that provide spatial evidence about the corresponding region, while spectral agreement with the partial shape and geometric regularization constrain the Hamiltonian potential beyond the observed locations. To further reduce the spatial ambiguity left by the sparse anchors, we introduce a reciprocal refinement mechanism that alternates between operator estimation and correspondence refinement. At each round, the eigenfunctions of the current operator guide correspondence estimation, while the induced support restricts the search to the candidate region. The refined correspondences, in turn, provide additional spatial evidence for updating the potential. This reciprocal process yields two complementary outputs (Fig.~\ref{fig:teaser}): the aggregated eigenfunction energy of the learned operator localizes the corresponding region, while the recovered map provides a reliable initialization for subsequent dense correspondence refinement. Our key contributions are as follows:

\begin{itemize}
    \item We propose an intrinsic neural-field parameterization of the Hamiltonian potential for learning a local spectral representation from sparse anchors and the intrinsic geometry of the partial shape.
    
    \item We introduce reciprocal operator--correspondence refinement to exploit the coupling between region localization and dense correspondence recovery.

    \item Extensive experiments demonstrate that the learned Hamiltonian operator supports both region localization and dense correspondence recovery from sparse anchors, while remaining robust to uniform scaling and rotation.
\end{itemize}

\section{Related Work}
\label{rw}

\paragraph{Partial Shape Correspondence.}
Functional maps~\citep{ovsjanikov2012functional} provide a compact spectral representation for shape correspondence. For partial matching, \citet{rodola2017partial} jointly estimate the corresponding region and a functional map, while FSPM~\citep{litany2017fully} constructs localized quasi-harmonic bases through joint approximate diagonalization. Learning-based methods use learned per-vertex features within functional-map frameworks, as in DPFM~\citep{attaiki2021dpfm} and ULRSSM~\citep{cao2023unsupervised}, or directly estimate pointwise correspondences from feature similarities~\citep{bracha2024unsupervised,bracha2024wormhole}. Another line of work~\citep{ehm2024geometrically,roetzer2025fast} uses discrete optimization to enforce geometric consistency in the recovered matchings. Our approach instead uses reliable sparse correspondences to guide the learning of a Hamiltonian potential, yielding a localized spectral representation for both region localization and dense correspondence recovery.

\paragraph{Hamiltonian Spectral Localization.}
Hamiltonian operators~\citep{choukroun2018hamiltonian} augment the Laplace--Beltrami operator (LBO) with a spatially varying potential, allowing low-energy eigenfunctions to concentrate in selected regions. \citet{rampini2019correspondence} exploit this property for correspondence-free region localization through Hamiltonian spectrum alignment. \citet{postolache2020parametric} further analyze the Hamiltonian--Dirichlet connection and exploit it for functional-map-based partial shape matching. Subsequent work~\citep{bensaid2022partial,bensaid2024multi} extends spectral region localization by aligning operator spectra under multiple intrinsic metrics. Building on this foundation, our method parameterizes the Hamiltonian potential as an intrinsic neural field and iteratively updates it using sparse anchors and refined correspondences.

\paragraph{Correspondence Refinement.}
Spectral correspondence refinement typically improves an initial map by alternating between spectral and pointwise representations. The original functional-map framework~\citep{ovsjanikov2012functional} introduces ICP-style refinement in spectral embedding space, while BCICP~\citep{ren2018continuous} further promotes bijectivity, continuity, and coverage during refinement. ZoomOut~\citep{zoomout} refines coarse or noisy initial maps by progressively increasing the spectral resolution. For partial matching, \citet{wu2020partial} combine Hamiltonian spectral alignment with ZoomOut-based iterative upsampling. More recently, NAM~\citep{vigano2025nam} generalizes functional-map refinement through a nonlinear neural representation and Neural ZoomOut. We adopt this approach to refine the correspondences obtained by NHO.

\section{Preliminaries}
\label{pre}

\paragraph{Laplace--Beltrami Operator.}
We model each shape as a compact, connected, 2-manifold $\mathcal{X}$, possibly with a smooth boundary $\partial\mathcal{X}$. We denote its interior by $\operatorname{int}(\mathcal{X})$. The positive semi-definite LBO $\Delta_{\mathcal{X}}$ generalizes the basic differential operator from Euclidean analysis to Riemannian manifolds. It admits an eigendecomposition
\begin{align}
    \Delta_{\mathcal{X}} \phi_i(x) &= \lambda_i \phi_i(x) \qquad x \in \operatorname{int}(\mathcal{X}) \label{eq:lbo_eigen}\\
    \phi_i(x) &= 0 \qquad x \in \partial\mathcal{X}, \label{eq:dirichlet_bc}
\end{align}
with homogeneous Dirichlet boundary conditions~(\plaineqref{eq:dirichlet_bc}), where $\{0\leq\lambda_1\leq\lambda_2\leq\cdots\}$ are the eigenvalues and $\phi_i$ are the corresponding eigenfunctions.

\paragraph{Hamiltonian Operator.}
As a theoretical foundation for our method, the Hamiltonian operator augments the LBO with a scalar potential function defined over the manifold~\citep{choukroun2018hamiltonian}. Given a non-negative potential $v:\mathcal{X}\rightarrow\mathbb{R}_{+}$, the Hamiltonian is defined as $H_{\mathcal{X}}=\Delta_{\mathcal{X}}+v$, acting on scalar functions as
\begin{equation}
    H_\mathcal{X}f = \Delta_{\mathcal{X}}f + vf, \label{eq:ha}
\end{equation}
where $vf$ denotes pointwise multiplication. For $v\equiv 0$, $H_{\mathcal{X}}$ reduces to $\Delta_{\mathcal{X}}$. On a compact manifold, $H_{\mathcal{X}}$ is self-adjoint on the same domain as $\Delta_{\mathcal{X}}$ and has a discrete spectrum. Its eigenpairs satisfy
\begin{equation}
H_{\mathcal{X}}\psi_i(x)=\mu_i\psi_i(x), \label{eq:hd}
\end{equation}
where, as before, the eigenvalues $\mu_i$ are listed in nondecreasing order, and the eigenfunctions $\psi_i$ are chosen to form an orthonormal basis of $L^2(\mathcal{X})$.

To illustrate the localization effect, consider a subdomain $\mathcal{R}\subset\mathcal{X}$ with smooth boundary and a finite step potential
\begin{equation}
    v_{\tau}(x)=\begin{cases}
        0 & x\in\mathcal{R}\\
        \tau & x\in\mathcal{X}\setminus\mathcal{R}
    \end{cases}
    \qquad \tau>0.
    \label{eq:tau}
\end{equation}
Here, $\tau$ denotes the potential barrier height outside $\mathcal{R}$. For sufficiently high barriers, low-energy Hamiltonian eigenfunctions concentrate in $\mathcal{R}$, although they generally do not vanish outside it~\citep{choukroun2018hamiltonian}. In this regime, the low-lying Hamiltonian eigenvalues approximate the corresponding Dirichlet eigenvalues of $\mathcal{R}$~\citep{postolache2020parametric}. This Hamiltonian--Dirichlet connection has been exploited for spectral region localization~\citep{rampini2019correspondence} and motivates our use of the learned Hamiltonian eigenfunctions for region localization and alignment with the Dirichlet eigenbasis of the partial shape.


\begin{figure}[t]
    \centering
    \includegraphics[width=\linewidth]{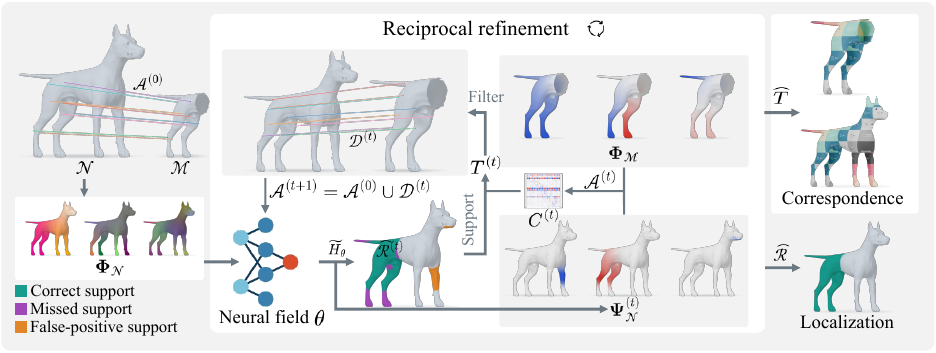}
    \caption{Overview of the NHO framework. Given a partial shape $\mathcal{M}$, a full shape $\mathcal{N}$, and sparse anchors $\mathcal{A}^{(0)}$, a neural field parameterizes the Hamiltonian potential on $\mathcal{N}$ using $\boldsymbol{\Phi}_{\mathcal{N}}$ as an intrinsic positional encoding. At round $t$, the Hamiltonian eigenspace induces the support $\widehat{\mathcal{R}}^{(t)}$ through aggregated energy and, together with $\boldsymbol{\Phi}_{\mathcal{M}}$ and $\mathcal{A}^{(t)}$, determines the alignment $C^{(t)}$. The support restricts the recovery of $T^{(t)}$, while geometrically reliable matches update the anchors and operator. The final support localizes the corresponding region, while the final-round map $\widehat{T}$ initializes dense correspondence refinement.}
    \label{fig:pipeline}
\end{figure}

\section{Methodology}
\label{method}
Fig.~\ref{fig:pipeline} provides an overview of our framework. We first introduce the problem and notation. Sec.~\ref{inhf} then presents the intrinsic neural Hamiltonian field (INHF), followed by neural Hamiltonian optimization, reciprocal refinement, and dense correspondence recovery in Sec.~\ref{rlc}.

\paragraph{Problem Formulation and Notation.}
Let $\mathcal{M}$ and $\mathcal{N}$ be connected partial and full triangle meshes with $n_{\mathcal{M}}$ and $n_{\mathcal{N}}$ vertices, respectively. We assume that $\mathcal{M}$ corresponds under a non-rigid deformation to an unknown region $\mathcal{R}\subseteq\mathcal{N}$. Given a sparse set of reliable anchors $\mathcal{A}^{(0)}=\{(p_\ell,q_\ell)\}_{\ell=1}^{m}$, where $p_\ell\in\mathcal{M}$ and $q_\ell\in\mathcal{N}$, our goal is to localize $\mathcal{R}$ and recover a dense map $T:\mathcal{M}\rightarrow\mathcal{N}$.

For vertices $p\in\mathcal{M}$ and $q\in\mathcal{N}$, let $a_p^{\mathcal{M}}$ and $a_q^{\mathcal{N}}$ denote the lumped vertex-area weights consistent with the Laplacian discretization. We denote the corresponding diagonal mass matrices by $A_{\mathcal{M}}$ and $A_{\mathcal{N}}$, and the resulting discrete surface areas by $\operatorname{Area}(\mathcal{M})$ and $\operatorname{Area}(\mathcal{N})$.

Let $\boldsymbol{\Phi}_{\mathcal{N}}^{j}$ denote the first $j$ mass-orthonormal LBO eigenfunctions of $\mathcal{N}$, with $\Lambda_{\mathcal{N}}^{j}$ the diagonal matrix of the corresponding eigenvalues. On $\mathcal{M}$, we impose Dirichlet boundary conditions and denote its first $j$ mass-orthonormal eigenfunctions by $\boldsymbol{\Phi}_{\mathcal{M}}^{j}$, with positive eigenvalues $\{\lambda_i^{\mathcal{M}}\}_{i=1}^{j}$.

\subsection{Intrinsic Neural Hamiltonian Field}
\label{inhf}

Building on the Hamiltonian--Dirichlet connection, we formulate region localization as spectral fitting.
When $\mathcal{M}$ and $\mathcal{R}$ are approximately isometric and share a common scale, aligning the low-frequency Hamiltonian spectrum on $\mathcal{N}$ with the Dirichlet spectrum of $\mathcal{M}$ encourages the low-potential region to recover $\mathcal{R}$. However, spectrum-based potential estimation is non-convex and does not uniquely determine the location of the support, making it sensitive to initialization~\citep{rampini2019correspondence}. In our setting, sparse anchors provide additional spatial evidence, while the area of $\mathcal{M}$ serves as a soft prior on the size of $\mathcal{R}$. These observations motivate a neural parameterization that allows spatial and spectral constraints to jointly guide operator estimation.

Following \citet{bensaid2023partial}, we use $\boldsymbol{\Phi}_{\mathcal{N}}^r$ as an intrinsic positional encoding. A single neural network $g_\theta:\mathbb{R}^{r}\rightarrow\mathbb{R}$ is applied row-wise to this encoding, with parameters shared across vertices, yielding an unconstrained scalar field $w_\theta\in\mathbb{R}^{n_{\mathcal{N}}}$:
\begin{equation}
    w_\theta= g_\theta\left(\boldsymbol{\Phi}_{\mathcal{N}}^{r}\right).
    \label{eq:neural_field}
\end{equation}
To obtain a differentiable representation of the candidate region, we transform $w_\theta$ into a soft region indicator:
\begin{equation}
    s_\theta= \frac{\mathbf{1}-\tanh(w_\theta)}{2},
    \label{eq:soft_region}
\end{equation}
where $\mathbf{1}$ is the all-ones vector and $\tanh$ is applied elementwise. We then define the bounded potential~\citep{rampini2019correspondence} and neural Hamiltonian:
\begin{align}
    v_\theta
    &=\tau(\mathbf{1}-s_\theta),
    \label{eq:neural_potential}\\
    H_{\mathcal{N},\theta}
    &=\Delta_{\mathcal{N}}+\operatorname{diag}(v_\theta).
    \label{eq:neural_hamiltonian}
\end{align}
High region scores correspond to low potential, with $s_\theta\in(0,1)^{n_{\mathcal{N}}}$ and $v_\theta\in(0,\tau)^{n_{\mathcal{N}}}$. As $s_\theta$ approaches a binary indicator, $v_\theta$ approaches the step potential in Eq.~\ref{eq:tau}. We denote the first $k$ Hamiltonian eigenpairs by $\{(\mu_i,\psi_i)\}_{i=1}^{k}$ and collect the eigenfunctions in $\boldsymbol{\Psi}_\mathcal{N}^k$, leaving their dependence on $\theta$ implicit. The field thus links the candidate region to a learned Hamiltonian spectral representation.

\subsection{Reciprocal Localization and Correspondence}
\label{rlc}

With the neural Hamiltonian representation defined above, we first estimate its parameters using sparse spatial evidence and the intrinsic geometry of $\mathcal{M}$.

\paragraph{Neural Hamiltonian optimization.}
Each anchor identifies a vertex $q_\ell\in\mathcal{N}$, so we encourage high region scores at these locations:
\begin{equation}
    \mathcal{L}_{\mathrm{anc}}=-\frac{1}{m}\sum_{\ell=1}^{m}\log s_\theta(q_\ell).
    \label{eq:anchor_loss}
\end{equation}

Anchors constrain location but not region size. We therefore use $\operatorname{Area}(\mathcal{M})$ as a soft area prior. The soft area induced by $s_\theta$ is $\sum_{q\in\mathcal{N}}a_q^{\mathcal{N}}s_\theta(q)$, leading to
\begin{equation}
    \mathcal{L}_{\mathrm{area}}=\left(
    \frac{
        \sum_{q\in\mathcal{N}}
        a_q^{\mathcal{N}}s_\theta(q)
        -\operatorname{Area}(\mathcal{M})
    }{
        \operatorname{Area}(\mathcal{N})}\right)^2.
    \label{eq:area_loss}
\end{equation}

Area alone does not enforce connectivity. We regularize the superlevel filtration of $s_\theta$ using zero-dimensional persistent homology~\citep{edelsbrunner2002topological,hu2019topology}:
\begin{equation}
    \mathcal{L}_{\mathrm{topo}}
    =\sum_{(b,d)\in\mathcal{P}_0(s_\theta)}
    \bigl(s_\theta(b)-s_\theta(d)\bigr)^2,
    \label{eq:topology_loss}
\end{equation}
where $\mathcal{P}_0(s_\theta)$ contains finite persistence pairs indexed by vertices $(b,d)$ attaining their birth and death values. Penalizing these pairs, excluding the unique essential component, encourages a single connected region.

These spatial terms alone do not ensure intrinsic compatibility with $\mathcal{M}$. Motivated by the Hamiltonian--Dirichlet connection~\citep{postolache2020parametric}, we encourage agreement with its first $k$ Dirichlet eigenvalues. We set the barrier above the target spectral range~\citep{rampini2019correspondence}:
\begin{equation}
    \tau=\beta\lambda_k^{\mathcal{M}},\qquad \beta>1.
    \label{eq:potential_height}
\end{equation}
The spectral loss is
\begin{equation}
    \mathcal{L}_{\mathrm{spec}}=
    \frac{1}{k}\sum_{i=1}^{k}
    \log\left(
        1+\left(
        \frac{\mu_i-\lambda_i^{\mathcal{M}}}
             {\lambda_i^{\mathcal{M}}}\right)^2\right).
    \label{eq:spectral_loss}
\end{equation}
Relative normalization balances errors across eigenvalue magnitudes, while the logarithm reduces the influence of large residuals.

For efficiency, we approximate the Hamiltonian eigensystem in the subspace spanned by the first $K$ LBO eigenfunctions of $\mathcal{N}$, where $k\leq K\ll n_{\mathcal{N}}$. Using the same mass matrix, we form the reduced operator:
\begin{equation}
    \widetilde{H}_\theta=\Lambda_{\mathcal{N}}^K+(\boldsymbol{\Phi}_{\mathcal{N}}^K)^\top
    A_{\mathcal{N}}\operatorname{diag}(v_\theta)\boldsymbol{\Phi}_{\mathcal{N}}^K.
    \label{eq:reduced_hamiltonian}
\end{equation}
We solve the resulting $K\times K$ differentiable eigenproblem and lift the first $k$ eigenvectors back to the mesh through $\boldsymbol{\Phi}_{\mathcal{N}}^K$.

Finally, we minimize
\begin{equation}
    \begin{aligned}
    \mathcal{L}_{\mathrm{op}}(\theta;\mathcal{A})
    ={}&\alpha_1\mathcal{L}_{\mathrm{anc}}(\mathcal{A})
       +\alpha_2\mathcal{L}_{\mathrm{area}}
       +\alpha_3\mathcal{L}_{\mathrm{topo}}
       +\alpha_4\mathcal{L}_{\mathrm{spec}},
    \end{aligned}
    \label{eq:operator_objective}
\end{equation}
where $\alpha_1,\ldots,\alpha_4\geq0$ weight the four terms. Optimization with $\mathcal{A}^{(0)}$ yields the initial parameters $\theta^{(0)}$.

\paragraph{Reciprocal Refinement.}
The initial estimate may remain spatially ambiguous, as illustrated in App.~\ref{app:reciprocal_visualization}. We therefore alternate operator updates with support-restricted correspondence estimation, using geometrically consistent matches as additional spatial evidence.

A superscript $(t)$ denotes quantities at round $t$. We extract the current region from aggregated eigenfunction energy:
\begin{align}
    e^{(t)}(q)
    &=\frac{1}{k}\|\boldsymbol{\Psi}_\mathcal{N}^{(t)}(q,:)\|_2^2,
    \label{eq:hamiltonian_energy}\\
    \widehat{\mathcal{R}}^{(t)}
    &=\left\{q\in\mathcal{N}\mid
    e^{(t)}(q)>\eta\mu_1^{(t)}\right\},
    \label{eq:hamiltonian_support}
\end{align}
where $\eta>0$ is a fixed empirical threshold factor. The estimated region restricts the target domain for matching. Using $\mathcal{A}^{(t)}$, we align the spectral coordinates:
\begin{equation}
    C^{(t)}\in
    \underset{C\in\mathbb{R}^{k\times k}}{\arg\min}
    \sum_{(p,q)\in\mathcal{A}^{(t)}}
    \left\|\boldsymbol{\Psi}_\mathcal{N}^{(t)}(q,:)C
    -\boldsymbol{\Phi}_{\mathcal{M}}^k(p,:)\right\|_2^2.
    \label{eq:reciprocal_alignment}
\end{equation}
We select the minimum-Frobenius-norm least-squares solution using a pseudoinverse. Restricted nearest-neighbor search then gives an intermediate map:
\begin{equation}
    T^{(t)}(p)
    =\underset{q\in\widehat{\mathcal{R}}^{(t)}}{\arg\min}
    \left\|\boldsymbol{\Phi}_{\mathcal{M}}^k(p,:)
    -\boldsymbol{\Psi}_\mathcal{N}^{(t)}(q,:)C^{(t)}\right\|_2^2.
    \label{eq:intermediate_correspondence}
\end{equation}
To limit unreliable feedback, we retain matches whose local mapping distortion (LMD) falls below a fixed threshold~\citep{xiang2021dual}. We merge them with the fixed original anchors, resolving conflicts among additional matches in favor of lower-LMD pairs. Denoting the retained additional pairs by $\mathcal{D}^{(t)}$, we update
\begin{equation}
    \mathcal{A}^{(t+1)}=\mathcal{A}^{(0)}\cup\mathcal{D}^{(t)},
    \label{eq:anchor_update}
\end{equation}
and minimize $\mathcal{L}_{\mathrm{op}}(\theta;\mathcal{A}^{(t+1)})$, warm-starting from $\theta^{(t)}$. The additional pairs are reselected at each round to adapt the spatial constraints on the operator.

\paragraph{Region Localization and Correspondence Recovery.}
After reciprocal refinement, the low-frequency Hamiltonian eigenfunctions exhibit stronger spatial concentration within the corresponding region, as visualized in App.~\ref{app:reciprocal_visualization}. We then freeze the operator and obtain the final region estimate $\widehat{\mathcal{R}}$ using Eq.~\ref{eq:hamiltonian_support}. For dense correspondence recovery, we denote the final-round map obtained by the restricted nearest-neighbor search in Eq.~\ref{eq:intermediate_correspondence} as $\widehat{T}$. We then refine $\widehat{T}$ using NAM-based Neural ZoomOut~\citep{vigano2025nam} to obtain the final correspondence $T$.



\section{Experiments}
\label{sec:experiments}

\begin{table}[!t]
\caption{Quantitative comparison of region localization on CUTS'24. All metrics are surface-area-weighted percentages; higher is better. Best in bold; second-best underlined.}
\label{tab:cuts24-localization}
\centering
\setlength{\tabcolsep}{3pt}
\renewcommand{\arraystretch}{1.15}
\resizebox{\linewidth}{!}{%
\begin{tabular}{l*{8}{r}@{\hspace{10pt}}*{8}{r}}
\toprule[1.2pt]
& \multicolumn{8}{c}{Default}
& \multicolumn{8}{c}{Normalized}
\\
\cmidrule[0.5pt](l{3pt}r{6pt}){2-9}
\cmidrule[0.5pt](l{6pt}r{3pt}){10-17}
& \multicolumn{4}{c}{Unrotated}
& \multicolumn{4}{c}{Rotated}
& \multicolumn{4}{c}{Unrotated}
& \multicolumn{4}{c}{Rotated}
\\
\cmidrule(lr){2-5}
\cmidrule(lr){6-9}
\cmidrule(lr){10-13}
\cmidrule(lr){14-17}
Method
& IoU & Precision & Recall & F1-score
& IoU & Precision & Recall & F1-score
& IoU & Precision & Recall & F1-score
& IoU & Precision & Recall & F1-score
\\
\midrule[0.6pt]
PFM
& \underline{71.88} & \underline{78.40}
& \underline{85.27} & \underline{81.60}
& \underline{72.22} & \underline{78.60}
& \underline{85.50} & \underline{81.82}
& 50.58 & 59.88 & 67.61 & 63.28
& 49.56 & 58.88 & 66.42 & 62.22
\\
FSPM
& 51.59 & 57.17 & 80.02 & 66.28
& 51.56 & 57.21 & 79.84 & 66.25
& 50.52 & 56.38 & 79.32 & 65.47
& 50.39 & 56.35 & 79.30 & 65.46
\\
Hamiltonian
& 50.73 & 60.34 & 62.37 & 61.02
& 50.38 & 60.02 & 61.58 & 60.48
& 52.43 & 62.04 & 64.40 & 62.85
& 52.96 & 62.69 & 65.37 & 63.63
\\
DPFM
& 17.03 & 53.48 & 20.06 & 26.16
& 13.33 & 42.37 & 15.82 & 20.73
& 80.45 & \underline{85.63} & 93.24 & 88.86
& 37.65 & 62.76 & 46.01 & 51.12
\\
Piecewise Smooth
& 0.00 & 0.00 & 0.00 & 0.00
& 0.00 & 0.00 & 0.00 & 0.00
& 73.88 & 83.77 & 84.74 & 82.84
& 49.84 & 64.43 & 63.15 & 61.59
\\
EchoMatch
& 43.15 & 50.25 & 76.34 & 58.19
& 41.30 & 49.58
& 75.21 & 56.43
& \textbf{82.14} & \textbf{86.95}
& \underline{93.80} & \textbf{89.94}
& \textbf{76.69} & \textbf{86.18}
& \underline{87.65} & \textbf{86.21}
\\
\midrule[0.6pt]
Ours
& \textbf{75.24} & \textbf{79.43}
& \textbf{94.11} & \textbf{85.37}
& \textbf{75.23} & \textbf{79.38}
& \textbf{93.95} & \textbf{85.36}
& 75.15 & 80.31 & 92.18 & 85.13
& \underline{75.06} & \underline{80.16}
& \textbf{92.08} & \underline{85.02}
\\
Ours (DPFM)
& 44.39 & 53.67 & 64.66 & 58.06
& 22.51 & 31.80 & 36.97 & 33.65
& \underline{80.92} & 84.92
& \textbf{93.91} & \underline{89.11}
& 53.01 & 61.46 & 68.08 & 64.45
\\
\bottomrule[1.2pt]
\end{tabular}%
}

\vspace{5pt}

\caption{Mean geodesic correspondence error ($\times100$, lower is better) on four partial shape matching benchmarks under four input settings. Avg. denotes the unweighted average over the four input settings
within each dataset. Best in bold; second-best underlined.}
\label{tab:multidataset-geodesic}
\centering
\setlength{\tabcolsep}{2pt}
\renewcommand{\arraystretch}{1.15}
\resizebox{\linewidth}{!}{%
\begin{tabular}{
l
*{4}{r}@{\hspace{7pt}}r@{\hspace{10pt}}
*{4}{r}@{\hspace{7pt}}r@{\hspace{10pt}}
*{4}{r}@{\hspace{7pt}}r@{\hspace{10pt}}
*{4}{r}@{\hspace{7pt}}r
}
\toprule[1.2pt]
& \multicolumn{5}{c}{CUTS'24}
& \multicolumn{5}{c}{PFAUST-M}
& \multicolumn{5}{c}{PFAUST-H}
& \multicolumn{5}{c}{PFARM}
\\
\cmidrule[0.5pt](l{3pt}r{5pt}){2-6}
\cmidrule[0.5pt](l{5pt}r{5pt}){7-11}
\cmidrule[0.5pt](l{5pt}r{5pt}){12-16}
\cmidrule[0.5pt](l{5pt}r{3pt}){17-21}
& \multicolumn{2}{c}{Default}
& \multicolumn{2}{c}{Normalized}
& \multirow{2}{*}{Avg.}
& \multicolumn{2}{c}{Default}
& \multicolumn{2}{c}{Normalized}
& \multirow{2}{*}{Avg.}
& \multicolumn{2}{c}{Default}
& \multicolumn{2}{c}{Normalized}
& \multirow{2}{*}{Avg.}
& \multicolumn{2}{c}{Default}
& \multicolumn{2}{c}{Normalized}
& \multirow{2}{*}{Avg.}
\\
\cmidrule(lr){2-3}
\cmidrule(lr){4-5}
\cmidrule(lr){7-8}
\cmidrule(lr){9-10}
\cmidrule(lr){12-13}
\cmidrule(lr){14-15}
\cmidrule(lr){17-18}
\cmidrule(lr){19-20}
Method
& Unrotated & Rotated
& Unrotated & Rotated
& {}
& Unrotated & Rotated
& Unrotated & Rotated
& {}
& Unrotated & Rotated
& Unrotated & Rotated
& {}
& Unrotated & Rotated
& Unrotated & Rotated
& {}
\\
\midrule[0.6pt]
PFM
& \textbf{5.03} & \textbf{4.97}
& 24.27 & 26.15
& \underline{15.10}
& 50.07 & 50.41
& 50.07 & 49.93
& 50.12
& 44.84 & 45.48
& 45.43 & 44.93
& 45.17
& 59.38 & 59.12
& 60.06 & 59.80
& 59.59
\\
FSPM
& 19.33 & 19.22
& 19.97 & 19.99
& 19.63
& 49.93 & 50.09
& 50.04 & 50.06
& 50.03
& 45.89 & 45.76
& 45.78 & 45.65
& 45.77
& 63.20 & 63.38
& 62.79 & 62.69
& 63.02
\\
DPFM
& 67.67 & 72.84
& \underline{2.38} & 24.94
& 41.96
& 34.46 & 36.64
& 37.88 & 38.93
& 36.98
& 38.60 & 38.68
& 39.62 & 41.71
& 39.65
& 49.53 & 53.14
& 51.59 & 45.82
& 50.02
\\
DPFM + ZoomOut
& 52.59 & 61.23
& \textbf{2.17} & 24.22
& 35.05
& 36.30 & 37.16
& 40.34 & 40.52
& 38.58
& 37.95 & 39.70
& 41.10 & 42.52
& 40.32
& 47.38 & 53.11
& 49.93 & 45.56
& 49.00
\\
ULRSSM
& 39.92 & 54.93
& 2.85 & 29.11
& 31.70
& 35.33 & 36.57
& 35.35 & 36.83
& 36.02
& 42.01 & 42.34
& 42.63 & 43.25
& 42.56
& 52.92 & 48.93
& 54.12 & 48.02
& 51.00
\\
EchoMatch
& 26.53 & 28.25
& 3.15 & \textbf{4.95}
& 15.72
& \textbf{9.21} & \underline{14.80}
& \textbf{10.02} & \underline{14.51}
& \underline{12.14}
& \textbf{16.91} & \textbf{21.53}
& \textbf{17.80} & \textbf{22.64}
& \textbf{19.72}
& \underline{23.16} & \underline{26.70}
& \underline{24.62} & \underline{28.01}
& \underline{25.62}
\\
Wormhole
& 38.17 & 55.60
& 4.56 & 42.75
& 35.27
& 28.82 & 42.12
& 29.17 & 42.19
& 35.58
& 30.65 & 45.50
& 33.43 & 46.06
& 38.91
& 46.43 & 47.45
& 48.27 & 51.10
& 48.31
\\
\midrule[0.6pt]
Ours
& \underline{7.03} & \underline{7.17}
& 6.53 & \underline{6.58}
& \textbf{6.83}
& \underline{12.20} & \textbf{11.85}
& \underline{12.36} & \textbf{11.63}
& \textbf{12.01}
& 26.94 & \underline{27.91}
& \underline{27.60} & \underline{27.50}
& \underline{27.49}
& \textbf{10.87} & \textbf{10.30}
& \textbf{9.12} & \textbf{10.13}
& \textbf{10.11}
\\
Ours (DPFM)
& 31.99 & 49.69
& 3.29 & 26.45
& 27.86
& 23.13 & 35.27
& 20.77 & 35.47
& 28.66
& \underline{26.89} & 36.76
& 31.64 & 38.20
& 33.37
& 33.56 & 38.74
& 46.38 & 48.24
& 41.73
\\
\bottomrule[1.2pt]
\end{tabular}%
}
\vspace{-4pt}
\end{table}

\begin{figure}[!t]
\centering

\includegraphics[width=\linewidth]{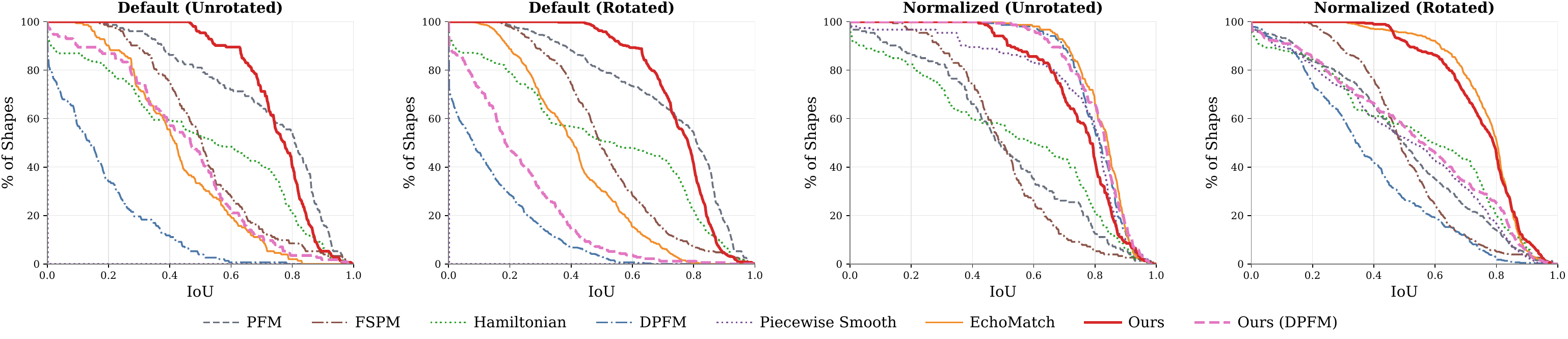}
\caption{IoU success curves for region localization on CUTS'24 under four input settings. Each point reports the percentage of shapes whose surface-area-weighted IoU exceeds the corresponding threshold.}
\label{fig:localization.iou}

\vspace{5pt}

\includegraphics[width=\linewidth]{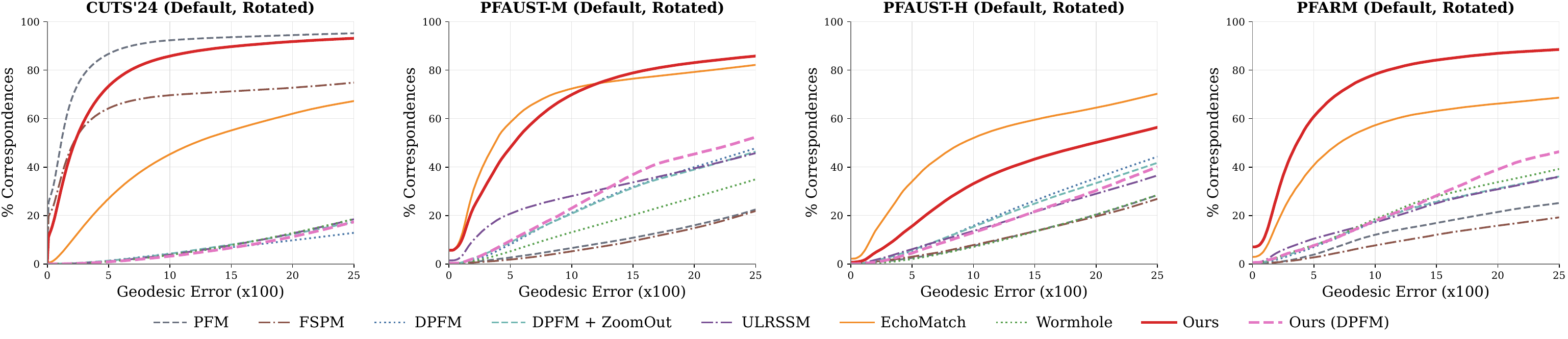}
\caption{PCK curves under the default-scale, rotated setting.}
\label{fig:geo.pck}

\vspace{-4pt}
\end{figure}

\paragraph{Implementation Details.}

We optimize NHO independently for each shape pair. By default, we use $m=25$ anchors selected by farthest-point sampling over the annotated partial vertices. Ours (DPFM) uses DPFM feature matches as initial anchors (App.~\ref{app:dpfm_anchors}); complete settings are provided in App.~\ref{app:implementation_details}.


\paragraph{Comparison Methods.}
We compare PFM~\citep{rodola2017partial}, FSPM~\citep{litany2017fully}, DPFM~\citep{attaiki2021dpfm}, and EchoMatch~\citep{xie2025echomatch} for both region localization and dense correspondence recovery.
For region localization, we additionally include the original correspondence-free Hamiltonian spectrum alignment method of \citet{rampini2019correspondence} and the piecewise-smooth localization method of \citet{bensaid2023partial}. For dense correspondence recovery, we further evaluate DPFM refined with ZoomOut~\citep{zoomout}, ULRSSM~\citep{cao2023unsupervised}, and Wormhole~\citep{bracha2024wormhole}.

\begin{figure}[t]
    \centering
    \includegraphics[width=\linewidth]{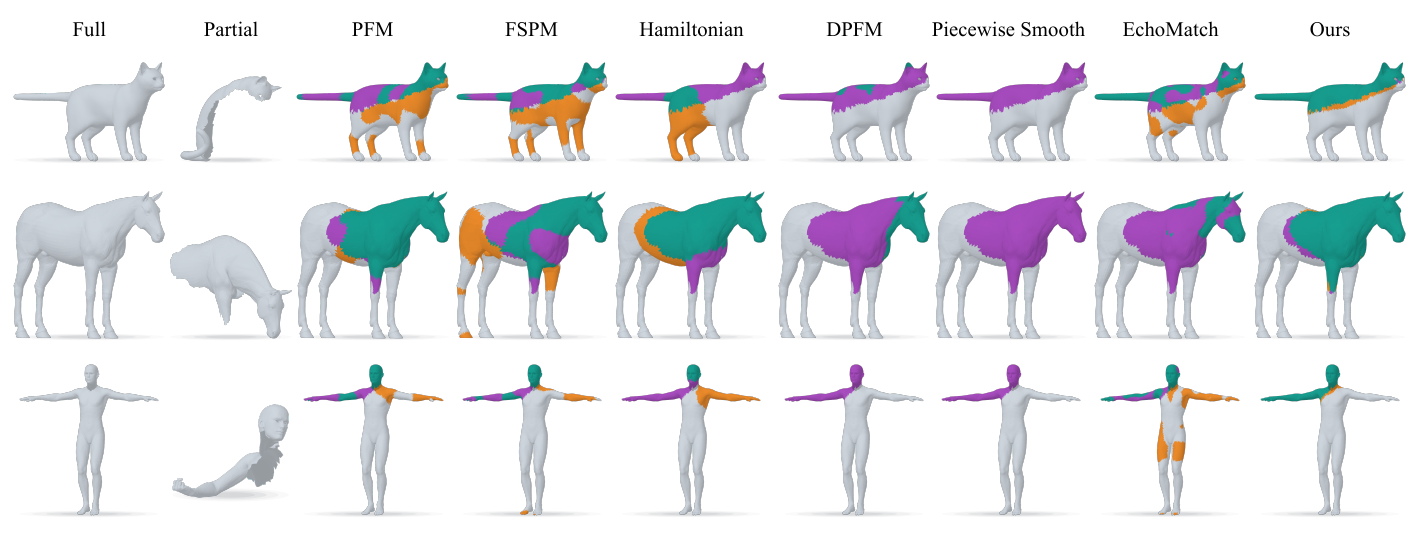}
    \caption{Qualitative comparison at the default scale under $\mathrm{SO}(3)$ rotations. Green, orange, and purple denote correct, false-positive, and missed support, respectively.}
    \label{fig:localization}
\end{figure}

\begin{figure}[t]
    \centering
    \includegraphics[width=\linewidth]{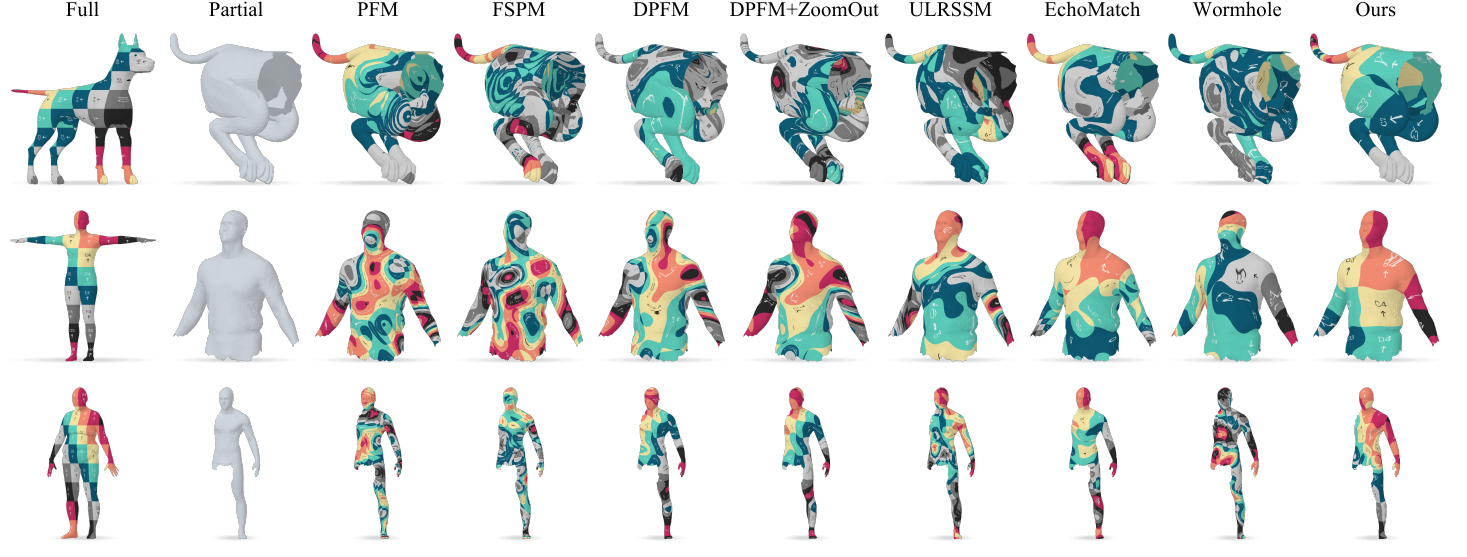}
    \caption{Qualitative comparison of partial-to-full correspondence. Our method yields more accurate correspondences than competing methods, even for challenging non-isometric shape pairs.}
    \label{fig:correspondence}
\end{figure}

\paragraph{Evaluation Datasets and Metrics.}
We evaluate region localization and dense correspondence on the leakage-reduced CUTS'24~\citep{ehm2024partial} split of SHREC'16 CUTS~\citep{lahner2016shrec}, PFAUST-M and PFAUST-H~\citep{bracha2024unsupervised}, and PFARM~\citep{attaiki2021dpfm}. The main paper reports localization results on CUTS'24 and correspondence results across all four datasets. Additional localization results on PFAUST-M, PFAUST-H, and PFARM and additional correspondence results are provided in Apps.~\ref{app:additional_localization} and~\ref{app:additional_correspondence}, respectively. To assess sensitivity to scale and orientation, we evaluate both default-scale and normalized inputs under unrotated and independently rotated $\mathrm{SO}(3)$ settings. Region localization is evaluated using surface-area-weighted intersection over union (IoU), precision, recall, and F1 score; IoU success curves report the fraction of test pairs exceeding each IoU threshold. Dense correspondence accuracy is reported using the mean geodesic error ($\times100$), with percentage of correct keypoints (PCK) curves illustrating performance across different error thresholds. Rotated results are averaged over three trials.

\subsection{Comparative Studies}

\paragraph{Region Localization.}

Tab.~\ref{tab:cuts24-localization} compares region localization on CUTS'24 under four scale--rotation settings. Traditional optimization-based methods remain sensitive to scale, whereas feature-learning methods perform well in the normalized regime used for feature training and degrade at the default scale. In contrast, NHO maintains IoU between $75.06$ and $75.24$ and F1 between $85.02$ and $85.37$, corresponding to variations of only $0.18$ and $0.35$ percentage points, respectively. The rotation stability follows from the intrinsic nature of the LBO and Hamiltonian operator. Under uniform rescaling, the area ratio remains unchanged, while the eigenvalues and the potential height $\tau=\beta\lambda_k^{\mathcal{M}}$ scale by the same inverse-square factor, leaving the relative spectral residual invariant. These properties explain why the optimization remains balanced across the tested transformations. The success curves in Fig.~\ref{fig:localization.iou} show that NHO is substantially more consistent across configurations. 


\paragraph{Dense Correspondence.}

Tab.~\ref{tab:multidataset-geodesic} reports dense correspondence accuracy across four benchmarks and four input settings. Across these settings, NHO achieves the lowest average error on CUTS'24, PFAUST-M, and PFARM, and the second-lowest average error on PFAUST-H. Its advantage lies primarily in consistency across transformations rather than peak performance in a single regime. For example, rotating normalized CUTS'24 increases the error of DPFM with ZoomOut from $2.17$ to $24.22$, whereas the error of NHO changes only from $6.53$ to $6.58$. This behavior is consistent with the learned operator: its localized support removes irrelevant candidates from the search domain, while its eigenfunctions provide intrinsic coordinates adapted to that support. Correspondence recovery therefore depends less on transformation-sensitive feature similarities. The PCK curves in Fig.~\ref{fig:geo.pck} show that this robustness extends across a range of error thresholds under the challenging default-scale, rotated setting. On normalized, unrotated CUTS'24, using reliable DPFM matches as anchors reduces the NHO error from $6.53$ to $3.29$, demonstrating that the operator can also incorporate stronger external correspondence evidence. 

\paragraph{Qualitative Results.}
Fig.~\ref{fig:localization} shows that NHO produces fewer missed and false-positive regions, while Fig.~\ref{fig:correspondence} demonstrates accurate dense correspondence, including on challenging non-isometric shape pairs. Additional qualitative and quantitative results are provided in Apps.~\ref{app:additional_localization} and~\ref{app:additional_correspondence}.


\subsection{Ablation Study}

\begin{wrapfigure}{R}{0.5\textwidth}
\vspace{-6pt}
\centering

\captionof{table}{Component ablations on normalized, unrotated CUTS'24. Localization metrics are area-weighted percentages; correspondence is measured by mean geodesic error ($\times100$) before and after external NAM refinement. The best result is shown in bold.}
\label{tab:ablation}

\vspace{4pt}

\setlength{\tabcolsep}{3pt}
\renewcommand{\arraystretch}{1.1}
\resizebox{\linewidth}{!}{%
\begin{tabular}{lcccccc}
\toprule[1.2pt]
&
\multicolumn{4}{c}{Region localization}
&
\multicolumn{2}{c}{Dense correspondence}
\\
\cmidrule(lr){2-5}
\cmidrule(lr){6-7}
Variant
& IoU $\uparrow$
& Precision $\uparrow$
& Recall $\uparrow$
& F1-score $\uparrow$
& \shortstack{Geo. before\\NAM $\downarrow$}
& \shortstack{Geo. after\\NAM $\downarrow$}
\\
\midrule[0.6pt]
w/o topology loss
& 74.45
& 80.09
& 91.28
& 84.69
& 8.71
& 6.85
\\
w/o reciprocal refinement
& 74.37
& 78.55
& \textbf{93.26}
& 84.61
& 9.23
& 6.97
\\
w/o Hamiltonian operator
& 55.48
& 68.08
& 74.20
& 70.94
& 12.69
& 9.84
\\
Full method
& \textbf{75.15}
& \textbf{80.31}
& 92.18
& \textbf{85.13}
& \textbf{8.37}
& \textbf{6.53}
\\
\bottomrule[1.2pt]
\end{tabular}%
}

\par\vspace{12pt}

\includegraphics[width=\linewidth]{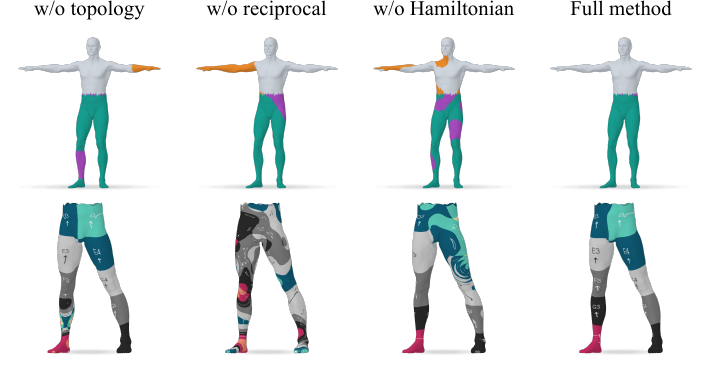}
\captionof{figure}{
Qualitative ablation results for region localization and dense
correspondence. Green, orange, and purple denote correct,
false-positive, and missed support, respectively.
}
\label{fig:ablation}

\vspace{-6pt}
\end{wrapfigure}

Tab.~\ref{tab:ablation} and Fig.~\ref{fig:ablation} examine the components of our default sparse-anchor configuration. Removing the topology loss has only a modest effect on the aggregate localization metrics, but produces spurious disconnected regions in the qualitative results. This indicates that the topology term mainly regularizes the spatial structure of the estimated support rather than substantially changing its overall overlap.

Reciprocal refinement improves support selectivity and correspondence accuracy. It increases precision from $78.55$ to $80.31$ and IoU from $74.37$ to $75.15$. Correspondence error decreases from $9.23$ to $8.37$ before NAM. These changes are consistent with reliable correspondence feedback reducing residual spatial ambiguity in an already plausible support, improving the balance between region coverage and the exclusion of distracting geometry.


Replacing the learned Hamiltonian operator with the global full-shape LBO reduces IoU from $75.15$ to $55.48$ and F1-score from $85.13$ to $70.94$, while increasing the final geodesic error from $6.53$ to $9.84$. This replacement produces the largest degradation among the evaluated ablations, highlighting the importance of a spectral representation adapted to the partial support.

Additional analyses are provided in App.~\ref{app:additional_analysis}.

\section{Conclusion}

We presented NHO, which learns a localized Hamiltonian operator on a full surface from sparse anchors and the intrinsic geometry of a partial shape. The learned operator localizes the corresponding region and supports dense matching, while geometrically reliable correspondences feed back into operator estimation. Experiments across multiple benchmarks demonstrate competitive performance and robustness to uniform scaling and rotation, while ablations validate the contributions of the neural Hamiltonian representation, topology regularization, and reciprocal refinement. These results show that our method can use sparse spatial evidence to learn a local operator for coupled region localization and correspondence. They further support the value of localized spectral representations for partial-to-full shape matching.


\clearpage

\subsection*{AI use statement}

Generative AI tools were used to draft and revise parts of the manuscript, improve readability, and assist with \LaTeX{} formatting. All AI-assisted text and formatting were reviewed by the authors. The authors take responsibility for the final content, including all claims and artifacts produced with the aid of generative AI.



\bibliography{iclr2027_conference}
\bibliographystyle{iclr2027_conference}

\clearpage

\appendix
\section*{Appendix}

\section{Methodological Comparison}
\label{app:method_comparison}

\begin{table}[h]
\caption{Comparison with representative partial-to-full correspondence methods. ``Anchor-source agnostic'' indicates that a method can accept sufficiently reliable sparse anchors without prescribing how they are obtained.}
\label{tab:method_comparison}
\centering
\setlength{\tabcolsep}{5pt}
\renewcommand{\arraystretch}{1.15}
\resizebox{\linewidth}{!}{%
\begin{tabular}{lccccc}
\toprule[1.2pt]
Method
& \shortstack{Region-localizing operator}
& \shortstack{Explicit region localization}
& \shortstack{Dense correspondence}
& \shortstack{Uniform-scale and rotation robustness}
& \shortstack{Anchor-source agnostic}
\\

\midrule[0.6pt]

PFM\citep{rodola2017partial}
& \xmark & \cmark & \cmark & \xmark & \xmark \\

FSPM\citep{litany2017fully}
& \xmark & \xmark & \cmark & \xmark & \xmark \\

Hamiltonian\citep{rampini2019correspondence}
& \cmark & \cmark & \xmark & \xmark & \xmark \\

DPFM\citep{attaiki2021dpfm}
& \xmark & \cmark & \cmark & \xmark & \xmark \\

Piecewise Smooth\citep{bensaid2023partial}
& \xmark & \cmark & \xmark & \xmark & \xmark \\

ULRSSM\citep{cao2023unsupervised}
& \xmark & \xmark & \cmark & \xmark & \xmark \\

RMR\citep{cao2024revisiting}
& \xmark & \xmark & \cmark & \xmark & \xmark \\

Synchronous Diffusion\citep{cao2024synchronous}
& \xmark & \xmark & \cmark & \xmark & \xmark \\

EchoMatch\citep{xie2025echomatch}
& \xmark & \cmark & \cmark & \xmark & \xmark \\

NFM\citep{cao2026hyper}
& \xmark & \xmark & \cmark & \xmark & \cmark \\

\midrule[0.6pt]

\textbf{Ours}
& \cmark & \cmark & \cmark & \cmark & \cmark \\

\bottomrule[1.2pt]
\end{tabular}%
}
\end{table}

Tab.~\ref{tab:method_comparison} summarizes the methodological differences between NHO and representative partial-to-full correspondence approaches. NHO retains the standard Hamiltonian construction \(H=\Delta+v\); its distinction lies in parameterizing the potential as an intrinsic neural field, estimating it from sparse anchors and partial-shape geometry, and using the resulting localized eigenspace for both support recovery and dense correspondence. Classical Hamiltonian spectrum alignment uses the operator primarily for region localization, whereas correspondence-oriented methods do not explicitly learn a region-localizing operator. NHO connects these roles through reciprocal operator--correspondence refinement.

\section{Visualization of Reciprocal Refinement}
\label{app:reciprocal_visualization}

\begin{figure}[h]
    \centering
    \includegraphics[width=\linewidth]{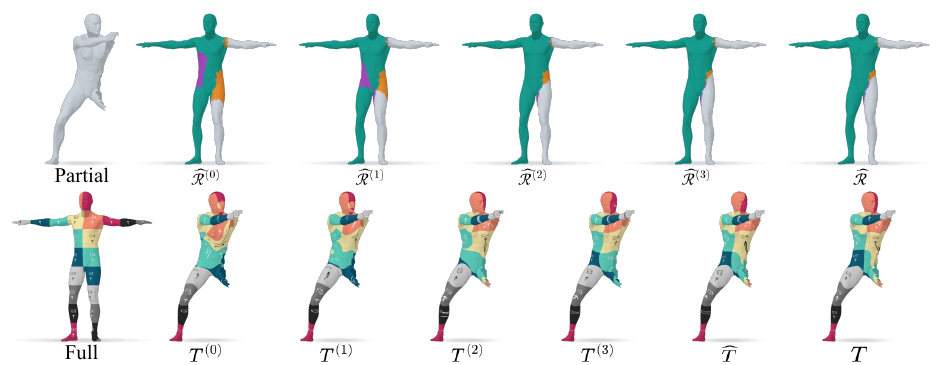}
    \caption{Visualization of reciprocal refinement. The estimated support and correspondence progressively improve across rounds. Green, orange, and purple denote correct, false-positive, and missed support, respectively.}
    \label{fig:refine}
\end{figure}

Fig.~\ref{fig:refine} visualizes the evolution of region localization and correspondence during reciprocal refinement. The initial estimate obtained from sparse anchors contains both false-positive and missed regions, accompanied by inaccurate correspondences. As geometrically reliable matches are fed back into operator estimation, the support becomes progressively more accurate and, in turn, provides a more reliable domain for subsequent correspondence recovery. The simultaneous reduction in support errors and improvement in correspondence consistency illustrate the reciprocal interaction between the two tasks.

\begin{figure}[t]
    \centering
    \includegraphics[width=\linewidth]{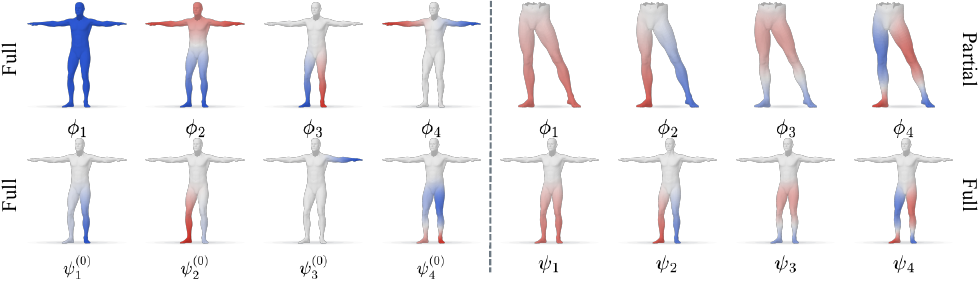}
    \caption{Evolution of the Hamiltonian eigenfunctions during reciprocal refinement. Top: the first four LBO eigenfunctions of the full shape (left) and Dirichlet eigenfunctions of the partial shape (right). Bottom: the anchor-aligned neural Hamiltonian eigenfunctions after initial operator estimation (left) and reciprocal refinement (right).}
    \label{fig:eigenfunction}
\end{figure}

Fig.~\ref{fig:eigenfunction} further shows how reciprocal refinement changes the learned spectral representation. For reference, the top row presents the LBO eigenfunctions of the full shape and the Dirichlet eigenfunctions of the partial shape. The bottom row compares the anchor-aligned Hamiltonian eigenfunctions before and after reciprocal refinement. Initially, the Hamiltonian modes remain weakly localized and differ from the partial Dirichlet modes. After refinement, they concentrate on the corresponding region and exhibit more consistent spatial patterns, explaining the improved localization and correspondence.

\section{DPFM-Based Anchor Initialization}
\label{app:dpfm_anchors}

Ours (DPFM) replaces the annotated anchors with correspondences derived from features extracted by a pretrained Siamese DiffusionNet~\citep{sharp2022diffusionnet,attaiki2021dpfm}. Let
$\mathbf{F}_{\mathcal{M}}\in\mathbb{R}^{n_{\mathcal{M}}\times d_f}$
and
$\mathbf{F}_{\mathcal{N}}\in\mathbb{R}^{n_{\mathcal{N}}\times d_f}$
denote the row-wise $\ell_2$-normalized feature matrices of the partial and full shapes, respectively. We first compute a dense feature-based nearest-neighbor map
\begin{equation}
\widetilde{T}_{\mathcal{M}\rightarrow\mathcal{N}}(p)
=
\underset{q\in\mathcal{N}}{\arg\max}\;
\left\langle
\mathbf{F}_{\mathcal{M}}(p),
\mathbf{F}_{\mathcal{N}}(q)
\right\rangle .
\label{eq:dpfm_direct_nn}
\end{equation}
Since the features are normalized, maximizing their inner product is equivalent to minimizing their Euclidean distance.

The initial operator optimization and spectral alignment impose different requirements on these matches. The anchor loss in Eq.~\ref{eq:anchor_loss} uses their target vertices as positive spatial evidence; an incorrect match can therefore directly distort the estimated support, making precision the primary concern. In contrast, estimating the alignment matrix in Eq.~\ref{eq:reciprocal_alignment} requires explicit correspondence pairs with sufficient coverage. We accordingly construct separate initial anchor sets for the two operations.

We measure the geometric consistency of
$\widetilde{T}_{\mathcal{M}\rightarrow\mathcal{N}}$
using local mapping distortion (LMD)~\citep{xiang2021dual}, computed over the $30$ nearest neighbors of each partial vertex under mesh-edge geodesic distance. The initial anchors for operator optimization are
\begin{equation}
\mathcal{A}_{H}^{(0)}
=
\left\{
\left(
p,\widetilde{T}_{\mathcal{M}\rightarrow\mathcal{N}}(p)
\right)
\;\middle|\;
\operatorname{LMD}_{\widetilde{T}}(p)<\delta_H
\right\},
\label{eq:dpfm_hamiltonian_anchors}
\end{equation}
where $\delta_H=0.42$. We use
$\mathcal{A}_{H}^{(0)}$
in
$\mathcal{L}_{\mathrm{op}}(\theta;\mathcal{A}_{H}^{(0)})$
to obtain the initial operator.

For spectral alignment, we additionally require bidirectional feature consistency. The reverse nearest-neighbor map is
\begin{equation}
\widetilde{T}_{\mathcal{N}\rightarrow\mathcal{M}}(q)
=
\underset{p\in\mathcal{M}}{\arg\max}\;
\left\langle
\mathbf{F}_{\mathcal{N}}(q),
\mathbf{F}_{\mathcal{M}}(p)
\right\rangle ,
\label{eq:dpfm_reverse_nn}
\end{equation}
from which we construct
\begin{equation}
\mathcal{A}_{C}^{(0)}
=
\left\{
(p,q)
\;\middle|\;
q=\widetilde{T}_{\mathcal{M}\rightarrow\mathcal{N}}(p),\;
p=\widetilde{T}_{\mathcal{N}\rightarrow\mathcal{M}}(q),\;
\operatorname{LMD}_{\widetilde{T}}(p)<\delta_C
\right\},
\label{eq:dpfm_alignment_anchors}
\end{equation}
where $\delta_C=1.5$. This set is used in Eq.~\ref{eq:reciprocal_alignment} to estimate the initial alignment matrix $C^{(0)}$.

\begin{table}[!t]
\caption{Quantitative comparison of region localization on PFAUST-M, PFAUST-H, and PFARM under four input settings. All metrics are surface-area-weighted percentages; higher is better. Best in bold; second-best underlined.}
\label{tab:additional-localization}
\centering
\setlength{\tabcolsep}{3pt}
\renewcommand{\arraystretch}{1.12}
\resizebox{\linewidth}{!}{%
\begin{tabular}{l*{8}{r}@{\hspace{10pt}}*{8}{r}}
\toprule[1.2pt]
& \multicolumn{8}{c}{Default}
& \multicolumn{8}{c}{Normalized}
\\
\cmidrule[0.5pt](l{3pt}r{6pt}){2-9}
\cmidrule[0.5pt](l{6pt}r{3pt}){10-17}
& \multicolumn{4}{c}{Unrotated}
& \multicolumn{4}{c}{Rotated}
& \multicolumn{4}{c}{Unrotated}
& \multicolumn{4}{c}{Rotated}
\\
\cmidrule(lr){2-5}
\cmidrule(lr){6-9}
\cmidrule(lr){10-13}
\cmidrule(lr){14-17}
Method
& IoU & Precision & Recall & F1-score
& IoU & Precision & Recall & F1-score
& IoU & Precision & Recall & F1-score
& IoU & Precision & Recall & F1-score
\\

\midrule[0.8pt]
\multicolumn{17}{l}{\textbf{PFAUST-M}}
\\
PFM
& 44.93 & 55.32 & 71.39 & 61.84
& 46.45 & 55.93 & 73.97 & 63.23
& 51.32 & 57.44 & \underline{83.14} & 67.50
& 50.14 & 56.87 & \underline{81.50} & 66.49
\\
FSPM
& 51.52 & 51.52 & \textbf{85.00} & 63.92
& 51.52 & 51.52 & \textbf{85.00} & 63.92
& 51.52 & 51.52 & \textbf{85.00} & 63.92
& 51.52 & 51.52 & \textbf{85.00} & 63.92
\\
Hamiltonian
& 51.48 & 60.66 & \underline{75.31} & 65.72
& \underline{51.72} & 61.01 & \underline{75.49} & \underline{66.05}
& 53.55 & 62.62 & 78.59 & 68.28
& \underline{52.96} & 62.41 & 77.48 & \underline{67.62}
\\
DPFM
& 13.95 & 68.74 & 15.01 & 24.02
& 19.20 & 61.35 & 22.05 & 31.77
& 19.75 & 66.73 & 21.94 & 32.69
& 17.82 & 57.22 & 20.54 & 29.94
\\
Piecewise Smooth
& 36.77 & 57.72 & 50.01 & 53.17
& 50.39 & 60.36 & 75.03 & 65.96
& 48.37 & 64.30 & 67.35 & 64.46
& 51.76 & 61.33 & 77.32 & 67.37
\\
EchoMatch
& \underline{51.69} & \textbf{81.42} & 58.43 & \underline{67.10}
& 41.46 & \textbf{77.42} & 46.74 & 57.11
& \underline{53.76} & \textbf{79.44} & 61.58 & \underline{68.73}
& 46.81 & \textbf{76.44} & 53.78 & 62.10
\\
\cmidrule(lr){1-17}
Ours
& \textbf{55.89} & \underline{70.89} & 71.58 & \textbf{70.77}
& \textbf{56.39} & \underline{71.90} & 71.95 & \textbf{71.53}
& \textbf{56.28} & \underline{71.74} & 71.87 & \textbf{71.40}
& \textbf{55.71} & \underline{71.09} & 71.50 & \textbf{70.90}
\\
Ours (DPFM)
& 47.47 & 61.81 & 65.81 & 63.54
& 48.19 & 60.17 & 69.71 & 64.33
& 52.97 & 64.38 & 73.41 & 68.50
& 46.64 & 58.58 & 67.30 & 62.45
\\

\midrule[0.8pt]
\multicolumn{17}{l}{\textbf{PFAUST-H}}
\\
PFM
& \underline{46.36} & 52.37 & \underline{80.53} & \textbf{63.18}
& \underline{46.09} & 52.30 & \underline{80.05} & \textbf{62.93}
& \textbf{49.28} & 53.37 & \textbf{86.58} & \textbf{65.91}
& \textbf{48.69} & 53.09 & \textbf{85.57} & \textbf{65.38}
\\
FSPM
& \textbf{46.75} & 46.75 & \textbf{85.00} & 60.23
& \textbf{46.75} & 46.75 & \textbf{85.00} & 60.23
& \underline{46.75} & 46.75 & \underline{85.00} & 60.23
& \underline{46.75} & 46.75 & \underline{85.00} & 60.23
\\
Hamiltonian
& 45.38 & 58.11 & 67.30 & \underline{61.54}
& 44.84 & 58.13 & 66.12 & \underline{60.97}
& 44.88 & 58.21 & 66.22 & 61.02
& 44.63 & 58.13 & 65.84 & \underline{60.79}
\\
DPFM
& 13.62 & \underline{61.75} & 14.87 & 23.70
& 18.52 & 54.37 & 22.00 & 30.80
& 19.25 & 58.67 & 22.13 & 32.03
& 17.62 & 51.84 & 20.98 & 29.66
\\
Piecewise Smooth
& 32.31 & 53.45 & 45.06 & 48.17
& 43.49 & 55.02 & 67.91 & 59.83
& 36.30 & 56.96 & 50.61 & 52.15
& 43.09 & 55.52 & 67.24 & 59.48
\\
EchoMatch
& 37.04 & \textbf{77.44} & 41.98 & 52.95
& 32.85 & \textbf{69.71} & 38.87 & 48.10
& 35.96 & \textbf{77.56} & 40.30 & 51.81
& 31.80 & \textbf{69.72} & 37.13 & 47.02
\\
\cmidrule(lr){1-17}
Ours
& 39.55 & 61.12 & 52.27 & 56.07
& 39.43 & \underline{61.09} & 52.59 & 56.14
& 40.20 & \underline{61.64} & 53.50 & 56.90
& 40.53 & \underline{61.76} & 53.89 & 57.15
\\
Ours (DPFM)
& 43.00 & 58.93 & 61.11 & 59.82
& 38.69 & 54.02 & 57.49 & 55.43
& 44.71 & 60.21 & 63.36 & \underline{61.54}
& 43.05 & 58.34 & 61.82 & 59.81
\\

\midrule[0.8pt]
\multicolumn{17}{l}{\textbf{PFARM}}
\\
PFM
& 15.21 & 33.12 & 20.80 & 24.62
& 14.32 & 32.31 & 19.65 & 23.63
& 13.70 & 31.58 & 18.42 & 22.57
& 14.33 & 32.22 & 19.40 & 23.49
\\
FSPM
& 20.04 & 20.04 & 36.00 & 25.35
& 19.03 & 20.38 & 34.36 & 24.75
& 22.64 & 25.20 & 39.50 & 29.50
& 21.98 & 24.67 & 37.95 & 28.39
\\
Hamiltonian
& 37.18 & 46.27 & 46.71 & 45.64
& 37.64 & 47.89 & 45.24 & 45.77
& 30.59 & 40.25 & 38.85 & 38.86
& 33.10 & 42.91 & 39.91 & 40.74
\\
DPFM
& 14.09 & 35.31 & 22.97 & 24.17
& 13.29 & 32.37 & 22.12 & 22.77
& 16.48 & 37.35 & 27.31 & 27.59
& 17.24 & 38.32 & 26.77 & 27.97
\\
Piecewise Smooth
& 37.07 & 44.05 & \underline{65.48} & 49.29
& 35.24 & 40.07 & \underline{66.57} & 46.82
& 37.49 & 46.68 & \underline{66.17} & 50.11
& 36.89 & 41.81 & \underline{67.95} & 48.89
\\
EchoMatch
& \underline{48.76} & \textbf{72.73} & 59.33 & \underline{62.71}
& \underline{45.30} & \textbf{69.83} & 56.15 & \underline{59.21}
& \underline{47.32} & \textbf{73.40} & 53.46 & \underline{60.64}
& \underline{42.68} & \textbf{71.33} & 49.40 & \underline{56.56}
\\
\cmidrule(lr){1-17}
Ours
& \textbf{62.71} & \underline{68.92} & \textbf{90.84} & \textbf{74.80}
& \textbf{62.76} & \underline{68.83} & \textbf{90.78} & \textbf{74.73}
& \textbf{63.50} & \underline{69.36} & \textbf{91.55} & \textbf{75.62}
& \textbf{63.60} & \underline{69.34} & \textbf{91.72} & \textbf{75.56}
\\
Ours (DPFM)
& 36.36 & 46.51 & 58.29 & 51.37
& 42.60 & 52.98 & 61.03 & 56.32
& 26.81 & 37.79 & 42.89 & 39.98
& 37.78 & 50.29 & 55.76 & 52.64
\\
\bottomrule[1.2pt]
\end{tabular}%
}

\par\vspace{4pt}
\begin{minipage}{\linewidth}
\centering
\includegraphics[width=0.94\linewidth]
{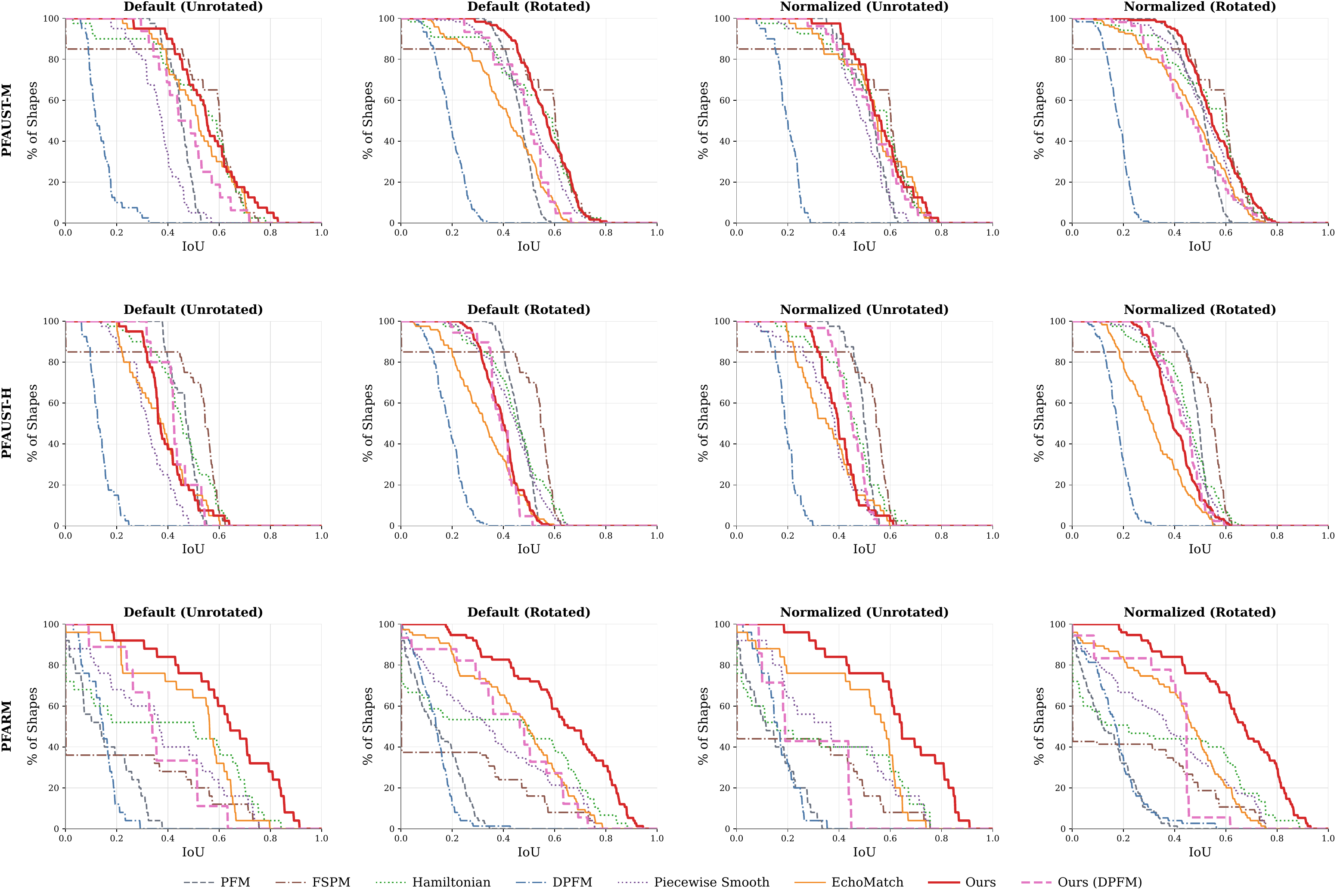}
\captionof{figure}{IoU success curves for region localization on PFAUST-M, PFAUST-H, and PFARM under four input settings.}
\label{fig:app_localization_curves}
\end{minipage}
\end{table}

Thus, $\mathcal{A}_{H}^{(0)}$ uses a strict LMD threshold to provide high-confidence spatial constraints for the anchor loss, whereas $\mathcal{A}_{C}^{(0)}$ combines mutual-nearest-neighbor consistency with a less restrictive threshold to retain sufficient pairs for spectral alignment. Subsequent operator--correspondence refinement proceeds as described in Sec.~\ref{rlc}. Neither initialization uses ground-truth region or correspondence annotations.

\section{Implementation Details}
\label{app:implementation_details}

\begin{figure}[t]
    \centering
    \includegraphics[width=\linewidth]{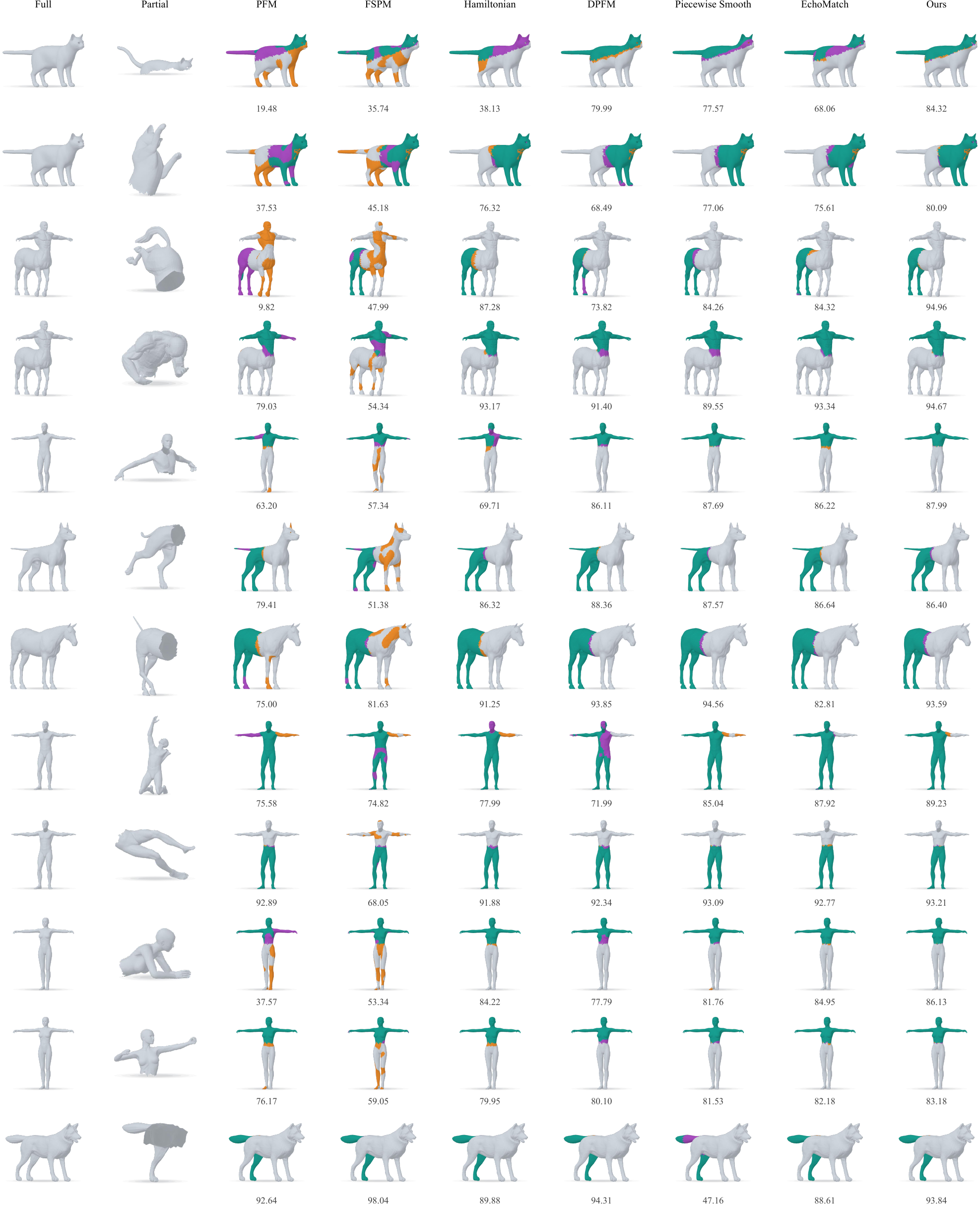}
    \caption{Additional qualitative comparisons of region localization. Numbers report the surface-area-weighted IoU for each prediction. Green, orange, and purple denote correctly localized, false-positive, and missed regions, respectively.}
    \label{fig:app_localization}
\end{figure}

The neural Hamiltonian field has three hidden layers of width $128$. We optimize the initial operator and each of four reciprocal updates for $100$ Adam steps with a learning rate of $10^{-2}$. We use $r=20$ LBO eigenfunctions for intrinsic positional encoding, $k=20$ Hamiltonian eigenpairs for spectral fitting and reciprocal alignment, and $K=100$ dimensions for Galerkin projection. We fix $(\alpha_1,\alpha_2,\alpha_3,\alpha_4)=(0.1,1.0,0.3,0.01)$, $\beta=50$, and $\eta=0.01$ across all datasets. LMD is computed over the $30$ nearest neighbors under mesh-edge geodesic distance, and matches below $0.42$ are retained.

After reciprocal refinement, we freeze the neural-field parameters and use the final-round map to initialize correspondence refinement. We independently apply per-coordinate RMS normalization to the Hamiltonian coordinates induced by the frozen operator and the partial Dirichlet coordinates. We then apply an adapted Neural ZoomOut~\citep{vigano2025nam}, increasing the spectral dimension from $20$ to $100$ in steps of $5$.

\section{Additional Localization and Correspondence Results}
\label{app:additional_results}

\begin{figure}[t]
    \centering
    \includegraphics[width=\linewidth]{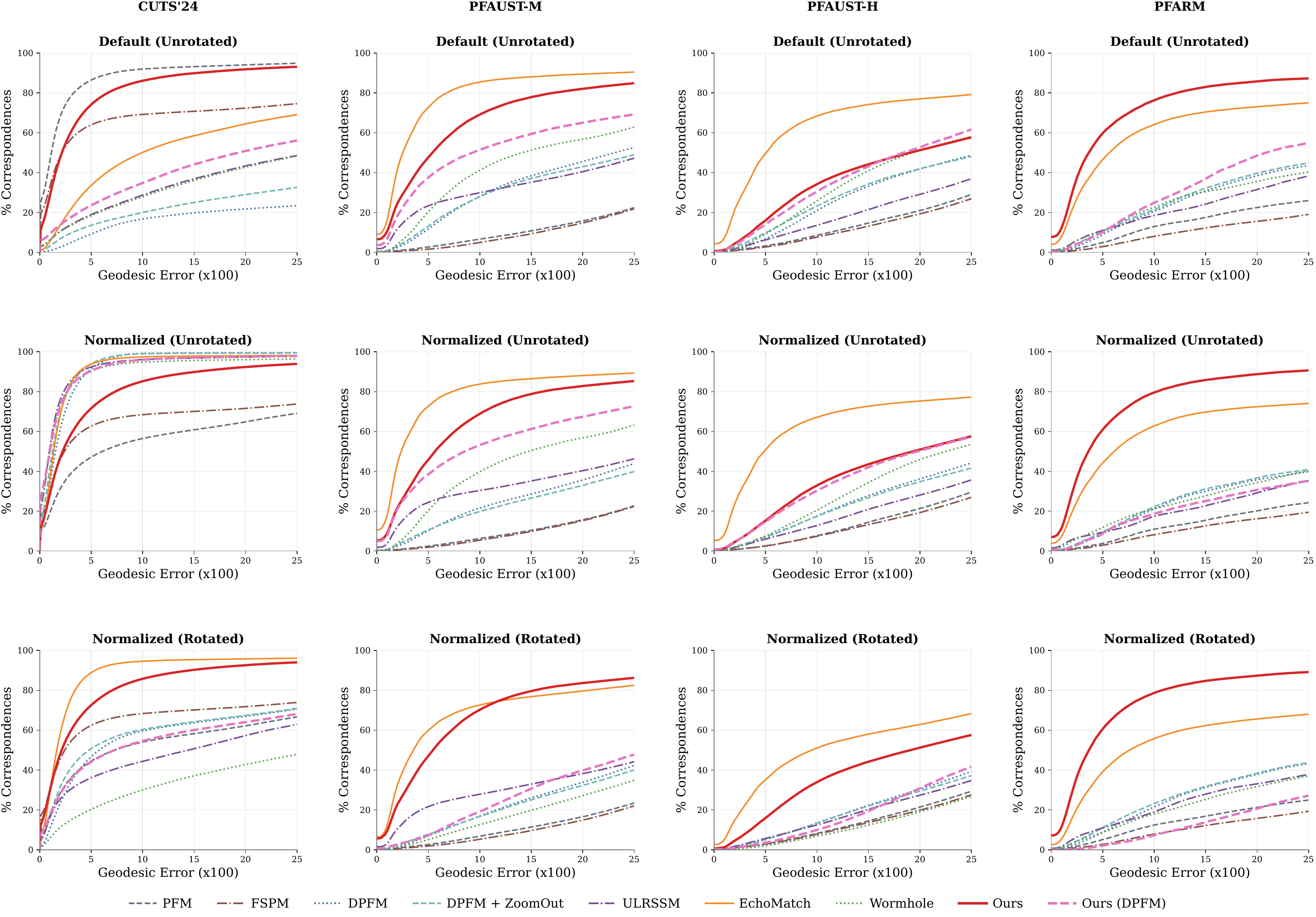}
    \caption{PCK curves on four benchmarks under the three input settings not shown in Fig.~\ref{fig:geo.pck}. Each point reports the percentage of correspondences whose geodesic error ($\times100$) does not exceed the threshold.}
    \label{fig:app_correspondence_curves}
\end{figure}

We provide additional quantitative and qualitative results for region localization and dense correspondence, complementing the comparisons in Sec.~\ref{sec:experiments}.

\subsection{Region Localization}
\label{app:additional_localization}

Tab.~\ref{tab:additional-localization} reports region localization results on PFAUST-M, PFAUST-H, and PFARM under four scale--rotation settings. On PFAUST-M, our default sparse-anchor configuration achieves the highest IoU and F1 score across all four settings. On PFARM, it also ranks first in both metrics. Its IoU ranges from $62.71\%$ to $63.60\%$, and its F1 score ranges from $74.73\%$ to $75.62\%$, while recall remains above $90\%$ throughout the four settings.

PFAUST-H is substantially more challenging. It contains a larger number of small missing regions, producing multiple additional boundary components and more pronounced topological changes~\citep{bracha2024unsupervised}. Although the Hamiltonian--Dirichlet connection remains applicable to such regions, their fine boundary structure is difficult to recover from a truncated low-frequency spectrum and a smoothly parameterized potential. Moreover, our zero-dimensional topological regularizer promotes connectivity but does not explicitly characterize the hole structure. The lower IoU and recall may therefore reflect the difficulty of resolving fine-scale hole boundaries using a truncated low-frequency representation.

\begin{figure}[t]
    \centering
    \includegraphics[width=\linewidth]{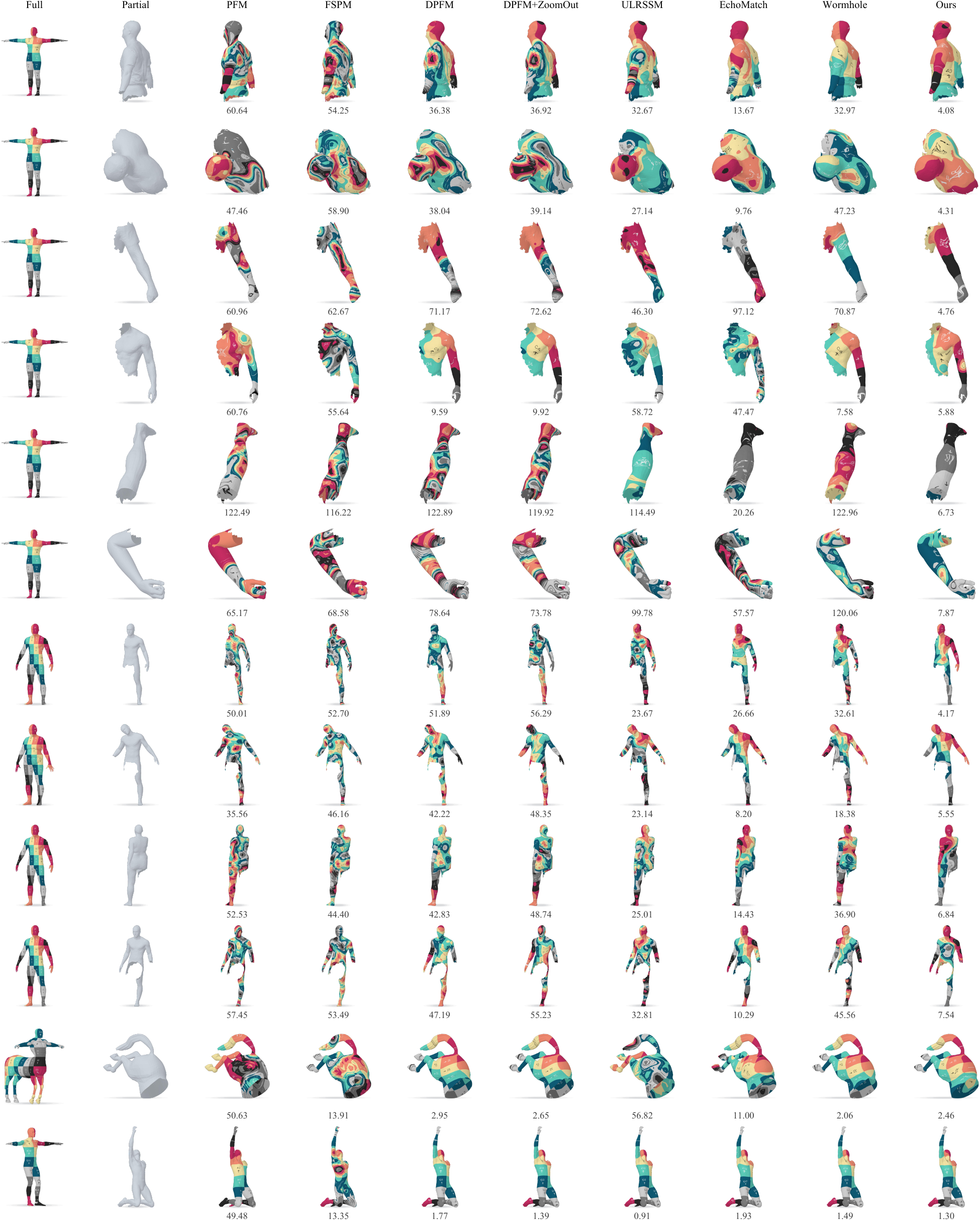}
    \caption{Additional qualitative comparisons of partial-to-full dense correspondence. Numbers report the mean geodesic error ($\times100$) for each case; lower is better.}
    \label{fig:app_correspondence}
\end{figure}

Despite the variation in absolute accuracy across datasets, sparse-anchor NHO remains stable under changes in scale and orientation. Its IoU varies by only $0.68$ percentage points on PFAUST-M, $1.10$ on PFAUST-H, and $0.89$ on PFARM across the four settings. The success curves in Fig.~\ref{fig:app_localization_curves} provide a distribution-level view of this consistency, while Fig.~\ref{fig:app_localization} presents additional qualitative comparisons of region localization.

\begin{table}[!t]
\centering
\setlength{\abovecaptionskip}{2pt}
\setlength{\belowcaptionskip}{2pt}
\setlength{\tabcolsep}{5pt}
\renewcommand{\arraystretch}{1.0}

\caption{Anchor sampling on normalized, unrotated CUTS'24.
Localization scores are area-weighted percentages; Geo. is the final
mean geodesic error ($\times100$). Best values are bold.}
\label{tab:anchor-sampling}
\begin{tabular}{lccccc}
\toprule[1.2pt]
& \multicolumn{4}{c}{Region localization}
& Dense correspondence \\
\cmidrule(lr){2-5}
\cmidrule(lr){6-6}
Sampling strategy
& IoU $\uparrow$ & Precision $\uparrow$ & Recall $\uparrow$
& F1-score $\uparrow$ & Geo. $\downarrow$ \\
\midrule[0.6pt]
Geodesic FPS
& \textbf{75.15} & \textbf{80.31} & 92.18
& \textbf{85.13} & \textbf{6.53} \\
Euclidean FPS
& 75.08 & 79.97 & \textbf{92.38} & 85.05 & 6.81 \\
Uniform random
& 72.38 & 78.67 & 89.73 & 83.21 & 7.92 \\
\bottomrule[1.2pt]
\end{tabular}

\par\vspace{2pt}
\includegraphics[width=0.90\linewidth]
{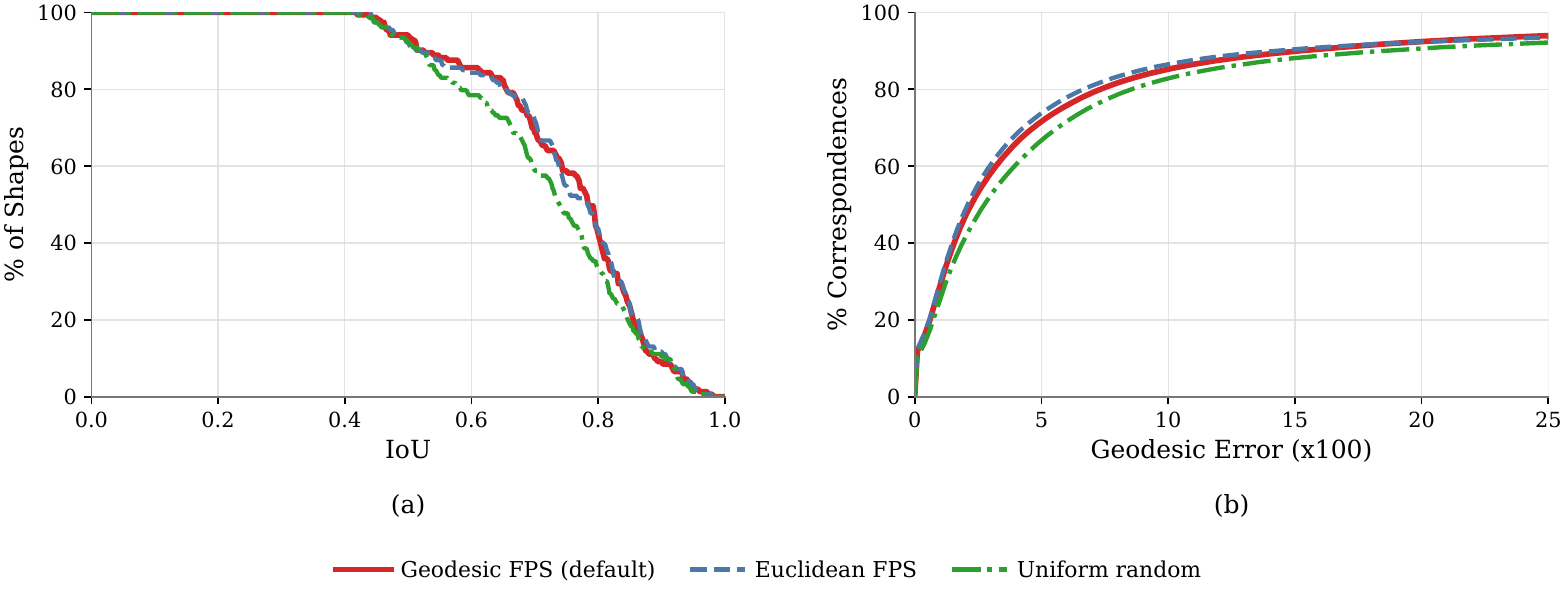}
\captionof{figure}{Anchor sampling: (a) IoU success and
(b) correspondence PCK curves.}
\label{fig:anchor-sampling}

\par\vspace{5pt}

\captionof{table}{Initial-anchor budget on normalized, unrotated
CUTS'24. Metrics follow Table~\ref{tab:anchor-sampling};
best values are bold.}
\label{tab:anchor-number}
\begin{tabular}{lccccc}
\toprule[1.2pt]
& \multicolumn{4}{c}{Region localization}
& Dense correspondence \\
\cmidrule(lr){2-5}
\cmidrule(lr){6-6}
$m$
& IoU $\uparrow$ & Precision $\uparrow$ & Recall $\uparrow$
& F1-score $\uparrow$ & Geo. $\downarrow$ \\
\midrule[0.6pt]
5   & 66.87 & 74.31 & 85.54 & 78.89 & 17.73 \\
15  & 72.78 & 78.68 & 90.26 & 83.38 & 11.43 \\
25  & 75.15 & 80.31 & 92.18 & 85.13 & 6.53 \\
50  & 75.49 & 80.60 & 92.14 & 85.29 & 4.75 \\
100 & \textbf{77.47} & \textbf{81.73} & \textbf{93.81}
    & \textbf{86.67} & \textbf{3.95} \\
\bottomrule[1.2pt]
\end{tabular}

\par\vspace{2pt}
\includegraphics[width=0.90\linewidth]
{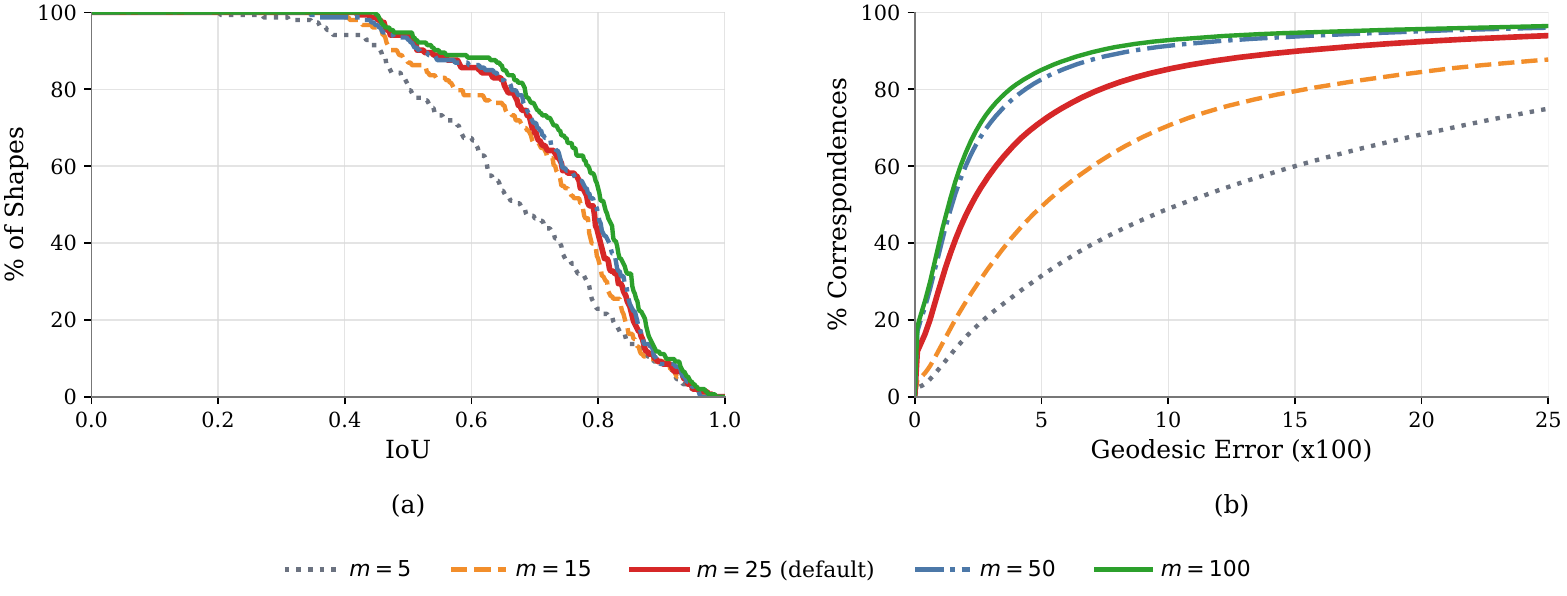}
\captionof{figure}{Initial-anchor budget: (a) IoU success and
(b) correspondence PCK curves.}
\label{fig:anchor-count}
\end{table}

\subsection{Dense Correspondence}
\label{app:additional_correspondence}

Tab.~\ref{tab:multidataset-geodesic} reports mean geodesic errors across four benchmarks and four input settings, while Figs.~\ref{fig:geo.pck} and~\ref{fig:app_correspondence_curves} provide the corresponding threshold-wise comparisons. The within-dataset error ranges of sparse-anchor NHO are $0.64$, $0.73$, $0.97$, and $1.75$ on the four benchmarks, respectively. This consistency indicates that the correspondence recovered from the learned operator is largely insensitive to the tested scale and orientation changes.

The PCK curves reveal additional behavior beyond the mean errors. On PFARM, NHO achieves higher PCK over most error thresholds in all four settings, showing that its advantage extends beyond the mean geodesic error to the overall distribution of correspondence accuracy. On PFAUST-M, EchoMatch is more accurate in the unrotated settings, whereas NHO becomes stronger after rotation: its error changes from $12.20$ to $11.85$ at the default scale and from $12.36$ to $11.63$ after normalization, while the corresponding EchoMatch errors increase from $9.21$ to $14.80$ and from $10.02$ to $14.51$. This comparison further highlights the rotation stability of the intrinsic operator representation.

Across the four benchmarks, sparse-anchor NHO records its lowest localization IoU and highest mean geodesic error on PFAUST-H. This shared degradation is consistent with the coupling between the two tasks: an inaccurate support may exclude valid target vertices or retain distracting regions, directly affecting the restricted correspondence search. Nevertheless, the correspondence error on PFAUST-H varies by only $0.97$ across the four settings, suggesting that its lower absolute accuracy is primarily associated with the more challenging partiality pattern discussed in Sec.~\ref{app:additional_localization}, rather than sensitivity to scale or rotation. Fig.~\ref{fig:app_correspondence} provides additional qualitative comparisons of the recovered dense correspondences.

\begin{table}[!t]
\caption{Robustness to mesh discretization on normalized,
unrotated CUTS'24. Localization metrics are surface-area-weighted
percentages; Geo. denotes the final mean geodesic error ($\times100$). Best values are shown in bold.}
\label{tab:discretization}
\centering
\setlength{\tabcolsep}{5pt}
\renewcommand{\arraystretch}{1.12}
\begin{tabular}{lccccc}
\toprule[1.2pt]
& \multicolumn{4}{c}{Region localization}
& \multicolumn{1}{c}{Dense correspondence} \\
\cmidrule(lr){2-5}
\cmidrule(lr){6-6}
Full faces retained
& IoU $\uparrow$
& Precision $\uparrow$
& Recall $\uparrow$
& F1-score $\uparrow$
& Geo. $\downarrow$ \\
\midrule[0.6pt]
50\% & 80.90 & 86.75 & 92.21 & 88.77 & \textbf{6.74} \\
20\% & 82.24 & \textbf{88.02} & 92.64 & 89.67 & 7.20 \\
10\% & \textbf{82.45} & 87.92 & \textbf{92.93}
     & \textbf{89.74} & 8.39 \\
5\%  & 80.44 & 86.16 & 92.17 & 88.52 & 10.56 \\
\bottomrule[1.2pt]
\end{tabular}

\par\vspace{8pt}

\begin{minipage}{\linewidth}
\centering
\includegraphics[width=\linewidth]{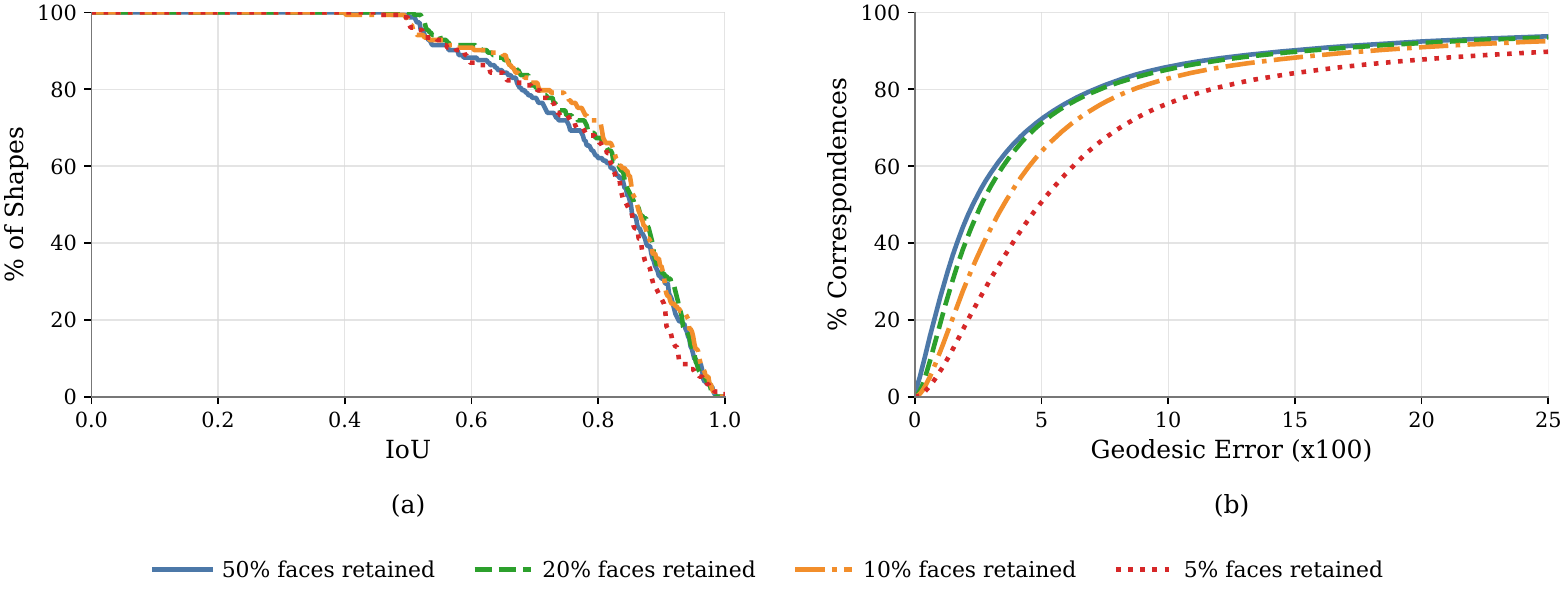}
\captionof{figure}{Robustness to mesh discretization on
normalized, unrotated CUTS'24. (a) IoU success curves for region
localization. (b) PCK curves for dense correspondence. Percentages
denote the retained full-shape faces.}
\label{fig:discretization_curves}
\end{minipage}
\end{table}

\section{Additional Analysis}
\label{app:additional_analysis}

\paragraph{Effect of Anchor Sampling.}
We fix the number of initial anchors at $m=25$ and compare the default geodesic FPS with Euclidean FPS and uniform random sampling over the annotated vertices of the partial shape. All other settings remain unchanged, and random sampling uses a fixed draw for each shape pair.

Tab.~\ref{tab:anchor-sampling} and Fig.~\ref{fig:anchor-sampling} show
that geodesic and Euclidean FPS yield nearly identical performance,
with a modest decline under uniform random sampling. These results
support NHO's robustness across the evaluated sampling strategies,
while the more uniform spatial coverage provided by FPS benefits
both localization and correspondence.

\paragraph{Effect of the Anchor Budget.}
We vary the number of initial anchors
$m\in\{5,15,25,50,100\}$ on normalized, unrotated CUTS'24 while
keeping all other settings fixed. When $m<k=20$, we set the spectral
correspondence dimension to $m$ to avoid an underdetermined
anchor-based alignment.

Tab.~\ref{tab:anchor-number} and Fig.~\ref{fig:anchor-count} show that
NHO remains effective with limited spatial evidence. Increasing $m$
from $5$ to the default budget of $25$ improves IoU from $66.87$ to
$75.15$ and reduces the final geodesic error from $17.73$ to $6.53$.
Beyond $25$ anchors, localization improves more gradually, whereas
correspondence continues to benefit from the additional constraints,
reaching an error of $3.95$ with $100$ anchors. We therefore use
$m=25$ uniformly across datasets as a sparse-anchor budget that
balances supervision and accuracy, rather than as the best-performing
anchor count.

\begin{figure}[t]
    \centering
    \includegraphics[width=\linewidth]{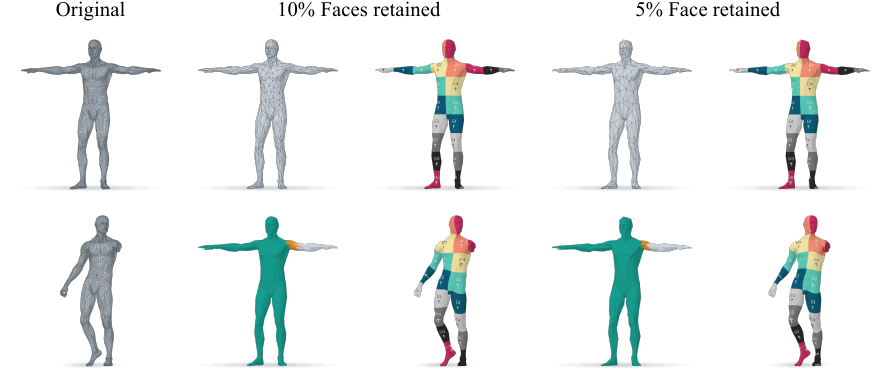}
    \caption{Qualitative results under severe discretization mismatch.
    The full meshes retain $10\%$ and $5\%$ of their original faces,
    while the partial meshes remain unchanged. Green, orange, and purple
    denote correct, false-positive, and missed support, respectively.}
    \label{fig:discretization_qualitative}
\end{figure}

\paragraph{Robustness to Mesh Discretization.}

We evaluate robustness to discretization mismatch by reducing the
resolution of the full meshes to $50\%$, $20\%$, $10\%$, and $5\%$
of their original faces while leaving the partial meshes unchanged.
The geometric operators are recomputed at each resolution, with all
other settings fixed.

Tab.~\ref{tab:discretization} and
Fig.~\ref{fig:discretization_curves} show that region localization
remains stable across the tested resolutions. IoU varies only from
$80.44\%$ to $82.45\%$, while F1 ranges from $88.52\%$ to $89.74\%$,
and the corresponding success curves largely overlap. This stability
is consistent with the low-frequency intrinsic formulation: when
large-scale geometry is preserved, the low-frequency LBO and
Hamiltonian subspaces remain comparatively stable under changes in
mesh connectivity and sampling density. Localization through
aggregated eigenfunction energy therefore does not rely on compatible
discretizations of the partial and full shapes.

Dense correspondence is more sensitive to aggressive simplification.
Reducing the retained full-shape faces from $50\%$ to $5\%$ increases
the final mean geodesic error from $6.74$ to $10.56$, with a
corresponding gradual decline in the PCK curves. Pointwise recovery
requires resolving local geometry and selecting individual target
vertices, both of which become less precise on coarser meshes.
Moreover, the higher-frequency coordinates used during correspondence
refinement are more sensitive to discretization than the low-frequency
energy used for localization. Nevertheless, the gradual degradation
shows that NHO retains useful correspondence accuracy even when the
full and partial shapes have substantially different resolutions.
Fig.~\ref{fig:discretization_qualitative} presents qualitative results
at the two most aggressive simplification levels.

\section{Computational Cost}

To reduce dependence on a dataset-specific training distribution, NHO
optimizes each shape pair independently rather than using amortized
feed-forward inference. This design improves adaptability across
datasets and input transformations at the cost of additional test-time
computation. The neural Hamiltonian field and the final
$100$-dimensional NAM stage contain $35.84$K and $52.34$K parameters,
respectively, with forward costs of $0.709$G and $1.040$G FLOPs for
$10{,}000$ vertices. Operator estimation requires $500$ Adam steps,
including the initial estimate and four reciprocal updates, while
multi-resolution correspondence refinement requires $3{,}400$ steps.
On an NVIDIA RTX 4090, the complete pipeline takes $143.15$ seconds per
shape pair on average, excluding cached spectral preprocessing.


\end{document}